\documentclass[twoside,11pt]{article}

\usepackage{geometry}
\usepackage[labelfont=bf]{caption}
\usepackage{fancyhdr}
\usepackage{float}
\usepackage{graphicx}
\usepackage{mathptmx}
\usepackage{latexsym}
\usepackage{colortbl}
\usepackage{dingbat}
\usepackage{pgfplots}
\usepackage{hyperref}
\usepackage{booktabs}
\usepackage{multirow}
\usepackage{fontawesome}
\usepackage{standalone}
\usepgfplotslibrary{groupplots}
\usepackage{tikz}
\usetikzlibrary{arrows,petri,topaths}
\usepackage{tkz-berge}
\usepackage[misc,geometry]{ifsym} 
\newcommand{\gc}[1]{\cellcolor{gray!25} #1}
\definecolor{nc}{HTML}{ff6f69}
\definecolor{pc}{HTML}{88d8b0}
\begin{document}

\title{\bf Current Limitations in Cyberbullying Detection: on Evaluation Criteria, Reproducibility, and Data Scarcity}

\author{Chris Emmery\thanks{CSAI, Tilburg University \hfill \Letter\ \ \texttt{cmry@pm.me}}\ \thanks{CLiPS, University of Antwerp}         \and
        Ben Verhoeven\footnotemark[2]       \and
        Guy De Pauw\footnotemark[2]           \and
        Gilles Jacobs\thanks{LT$^3$, Ghent University}        \and
        Cynthia Van Hee\footnotemark[3]        \and
        Els Lefever\footnotemark[3]            \and
        Bart Desmet\footnotemark[3]           \and
        V\'{e}ronique Hoste\footnotemark[3]    \and
        Walter Daelemans\footnotemark[2]
}

\date{}

\maketitle


\begin{abstract}

The detection of online cyberbullying has seen an increase in societal importance, popularity in research, and available open data. Nevertheless, while computational power and affordability of resources continue to increase, the access restrictions on high-quality data limit the applicability of state-of-the-art techniques. Consequently, much of the recent research uses small, heterogeneous datasets, without a thorough evaluation of applicability. In this paper, we further illustrate these issues, as we (i) evaluate many publicly available resources for this task and demonstrate difficulties with data collection. These predominantly yield small datasets that fail to capture the required complex social dynamics and impede direct comparison of progress. We (ii) conduct an extensive set of experiments that indicate a general lack of cross-domain generalization of classifiers trained on these sources, and openly provide this framework to replicate and extend our evaluation criteria. Finally, we (iii) present an effective crowdsourcing method: simulating real-life bullying scenarios in a lab setting generates plausible data that can be effectively used to enrich real data. This largely circumvents the restrictions on data that can be collected, and increases classifier performance. We believe these contributions can aid in improving the empirical practices of future research in the field.

\end{abstract}

\section{Introduction}

Learning to accurately classify rare phenomena within large feeds of data poses challenges for numerous applications of machine learning. The volume of data required for representative instances to be included is often resource-consuming, and limited access to such instances can severely impact the reliability of predictions. These limitations are particularly prevalent in applications dealing with sensitive social phenomena such as those found in the field of forensics: e.g., predicting acts of terrorism, detecting fraud, or uncovering sexually transgressive behavior. Their events are complex and require rich representations for effective detection. Conversely, online text, images, and meta-data capturing such interactions have commercial value for the platforms they are hosted on and are often off-limits to protect users' privacy.

An application affected by such limitations with increasing societal importance and growing interest over the last decade is that of cyberbullying detection. Not only is it sensitive, but the data is also inherently scarce in terms of public access. Most cyberbullying events are off-limits to the majority of researches, as they take place in private conversations. Fully capturing the social dynamics and complexity of these events requires much richer data than available to the research community up until now. Related to this, various issues with the operationalization of cyberbullying detection research were recently demonstrated by \cite{rosa2019automatic}, who share much of the same concerns as we will discuss in this work. While their work focuses on methodological rigor in prior research, we will focus on the core limitations of the domain and complexity of cyberbullying detection. Through an evaluation of the current advances on the task, we illustrate how the mentioned issues affect current research, particularly cross-domain. Finally, we demonstrate crowdsourcing in an experimental setting to potentially alleviate the task's data scarcity. First, however, we introduce the theoretical framing of cyberbullying and the task of automatically detecting such events.

\subsection{Cyberbullying} \label{sec:bul}

Asynchrony and optional anonymity are characteristic of online communication as we know it today; it heavily relies on the ability to communicate with people who are not physically present, and stimulates interaction with people outside of one's group of close friends through social networks \cite{Madden2013}. The rise of these networks brought various advantages to adolescents: studies show positive relationships between online communication and social connectedness \cite{Bessiere2008,Valkenburg2007}, and that self-disclosure on these networks benefits the quality of existing and newly developed relationships \cite{Steijn2013}. The popularity of social networks and instant messaging among children has resulted in this age group using devices that are connected to the Internet from increasingly younger ages \cite{Olafsson2013}, with $95\%$ of teens\footnote{Survey conducted in 2011 among 799 American teens. Black and Latino families were oversampled.} ages 12--17 online, of which $80\%$ are on social media \cite{Lenhart2011}. For them, however, the transition from social interaction predominantly taking place on the playground to being mediated through mobile devices \cite{Livingstone2011} has also moved negative communication to a platform where indirect and anonymous interaction has a window into homes.

A range of studies conducted by the Pew Research Center\footnote{\url{www.pewinternet.org}}, most notably \cite{Lenhart2011}, provides detailed insight into these developments. While $78\%$ of teens report positive outcomes from their social media interactions, $41\%$ have experienced at least some adverse outcomes, ranging from arguments, trouble with school and parents, physical fights and ending friendships. From $19\%$ bullied in the 12 months prior to the study, $8\%$ of all teens reported this was some form of cyberbullying. These numbers are comparable to other research \cite{Morgan2015,Kann2014} (7\% for grades 6--12, and 15\% grades 9--12 respectively). Bullying has for a while been regarded as a public health risk by numerous authorities \cite{Xu2012}, with depression, anxiety, low self-esteem, school absence, lower grades, and risk of self-medication as primary concerns.

The act of cyberbullying---other than being conducted online---shares the characteristics of traditional bullying: a power imbalance between the bully and victim \cite{Sharp2002}, the harm is intentional, repeated over time, and has a negative psychological effect on the victim \cite{DeHue2008}. With the Internet as a communication platform however, some additional aspects arise: location, time, and physical presence have become an irrelevant factor in the act. Accordingly, several categories unique to this form of bullying are defined \cite{Willard2007,Beran2008}: \emph{flaming} (sending rude or vulgar messages), \emph{outing} (posting private information or manipulated personal material of an individual without consent), \emph{harassment} (repeatedly sending offensive messages to a single person), \emph{exclusion} (from an online group), \emph{cyberstalking} (terrorizing through sending explicitly threatening and intimidating messages), \emph{denigration} (spreading online gossips), and \emph{impersonation}. Moreover, in addition to optional anonymity hiding the critical figures behind an act of cyberbullying, it could also obfuscate the number of actors (i.e., there might only be one even though it seems there are more). Cyberbullying acts can prove challenging to remove once published; messages or images might persist through sharing and be viewable by many (as is typical for hate pages), or available to a few (in group or direct conversations). Hence, it can be argued that any form of harassment has become more accessible and intrusive. This online nature has an advantage as well: in theory, platforms record these bullying instances. Therefore, an increasing number of researches are interested in the automatic detection (and prevention) of
cyberbullying.

\subsection{Detection and Task Complexity}

The task of cyberbullying detection can be broadly defined as the use of machine learning techniques to automatically classify text in messages on bullying content, or infer characteristic features based on higher-order information, such as user features or social network attributes. Bullying is most apparent in younger age groups through direct verbal outings \cite{Vaez2004}, and more subtle in older groups, mainly manifested in more complex social dynamics such as exclusion, sabotage, and gossip \cite{Privitera2009}. Therefore, the majority of work on the topic focuses on younger age groups, be it deliberately or given that the primary source for data is social media---which will likely result in these being highly present for some media \cite{Duggan2015}. Apart from the well-established challenges that language-use poses (e.g., ambiguity, sarcasm), two factors in the event add further linguistic complexity,  namely that of actor \emph{role} and associated \emph{context}. In contrast to tasks where adequate information is provided in the text of a single message alone, to completely map a cyberbullying event and pinpoint bully and victim implies some understanding of the dynamics between the involved actors and the concurrent textual interpretation.

\begin{figure*}[t] 
    \centering
    \begin{tikzpicture}[scale=1,transform shape]
  \Vertex[x=0.0,y=2.0]{AB}
  \Vertex[x=2.0,y=3.0]{B}
  \Vertex[x=5.0,y=2.0]{V}
  \Vertex[x=7.0,y=3.0]{S}
  \Vertex[x=7.0,y=2.0]{VF}
  \Vertex[x=7.0,y=4.0, style=dotted]{R}
  \Vertex[x=0.0,y=4.0]{BY}
  \Vertex[x=5.0,y=4.0, style=dotted]{A}
  \Vertex[x=0.0,y=3.0]{BF}
  \tikzstyle{LabelStyle}=[fill=white,sloped]

  \Edge[label=$-$](AB)(V)
  \Edge[label=$-$, lw=2pt](B)(V)
  \Edge[label=$-$, lw=2pt](A)(B)
  \Edge[label=$+$](AB)(B)
  \Edge[label=$+$](S)(V)
  \Edge[label=$-$](S)(B)
  \Edge[label=$+$, style=dotted](BF)(B)
  \Edge[label=$+$, style=dotted](VF)(V)
\end{tikzpicture}
    \caption{Role graph of a bullying event. Each vertex represents an actor,
             labeled by their role in the event. Each edge indicates a stream of
             communication, labeled by whether this is positive ($+$) or negative
             ($-$) in nature, and its strength the frequency of interaction. The
             dotted vertices were added by \cite{Xu2012} to account for
             social-media-specific roles.}
    \label{fig:graph}
\end{figure*}
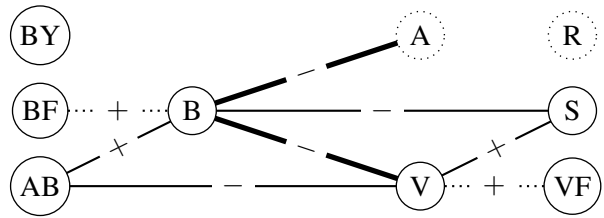

\paragraph{Roles} \label{part:roles}
Firstly, there is a commonly made distinction between several actors within a cyberbullying event. A naive role allocation includes a bully $B$, a victim $V$ and bystander $BY$, the latter of whom may or may not approve of the act of bullying. More nuanced models such as that of \cite{Xu2012} include the additional roles of reinforcer $BF$, assistant $AB$, defender $S$, reporter $R$, and accuser $A$. Different roles can be assigned to one person; for example, being bullied and reporting this---they are visualized in Figure~\ref{fig:graph}. Most importantly, all shown roles can be present in the span of one single thread on social media, as demonstrated in Table~\ref{tab:bul}. While some roles clearly show from frequent interaction with either a positive or negative sentiment ($B$, $V$, $A$), others might not be observable through any form of conversation ($R$, $BY$), are too subtle, or not distinguishable from other roles. 

\begin{table*}[t] \small 
    \centering
    \caption{Fictional example of a cyberbullying conversation. Lines represent sequential turns. Roles are noted as described on Page~\pageref{part:roles} (under the eponymous paragraph), if the message can be considered bullying by \checkmark, and types according to \cite{VanHee2015guide}.}
        \begin{tabular*}{\textwidth}{@{\extracolsep{\fill}} rllcr}
      \hline\noalign{\smallskip}
      Line & Role & Message & Bully & Type \\
      \noalign{\smallskip}\hline\noalign{\smallskip}
      1 & V    & me and my friends hanging out tonight! :)                 &   & neutral \\
      2 & B    & @V lol b*tch, you dont have any friends.. ur fake as sh*t & \checkmark & curse, insult \\
      3 & AB   & @B haha word, shes so sad                                 & \checkmark & encouragement \\
      4 & VF   & @V you know it girl                                       &   & \\
      5 & S    & @V dont listen to @B, were gonna have fun for sure!        &   & defense \\
      6 & V    & @B shut up @B!! nobody asked your opinion!!!!              &   & defense \\
      7 & A    & @B you are a f*cking bully, go outside or smt             &   & insult \\
      8 & B    & @V @S haha you all so dumb, just kill yourself already!   & \checkmark & insult, curse \\
      9 & A, R & @B shut up or ill report you                                &   & \\
     10 & B    & @A u gonna cry? go ahead, see what happens tomorrow!      & \checkmark & threat \\
      \noalign{\smallskip}\hline
  \end{tabular*}
    \label{tab:bul}
\end{table*}

\paragraph{Context} Secondly, the content of the messages has to be interpreted differently between these roles. While curse words can be a good indication of harassment, identification of a bully arguably requires more than these alone. Consider Table~\ref{tab:bul}: both $B$ and $A$ use insults (lines 7--8), the message of $V$ (line 6) might be considered as bullying in isolation, and having already determined $B$, the last sentence (line 10) can generally be regarded as a threat. In conclusion, the full scope of the task is complex; it could have a temporal-sequential character, would benefit from determining actors and their interactions, and then should have some sense of severity as well (e.g.\ distinguish bullying from teasing).

\subsection{Our Contributions}

Surprisingly, a significant amount of work on the task does not collect (or use) data that allows for the inference of such features (which we will further elaborate on in Section~\ref{sec:oos}). To confirm this, we reproduce part of the previous cyberbullying detection research on different sources. Predictions made by current automatic methods for cyberbullying classification are demonstrated not to reflect the above-described task complexity; we show performance drops across different training domains, and give insights into content feature importance and limitations. Additionally, we report on reproducibility issues in the current state-of-art work when subjected to our evaluation. To facilitate future reproduction, we will provide all code open-source, including dataset readers, experimental code, and qualitative analyses.\footnote{Available at \url{https://github.com/cmry/amica}.} Finally, we present a method to collect crowdsourced cyberbullying data in an experimental setting. It grants control over the size and richness of the data, does not invade privacy, nor rely on external parties to facilitate data access. Most importantly, we demonstrate that it successfully increases classifier performance. With this work, we provide suggestions on improving methodological rigor and hope to aid the community in a more realistic evaluation and implementation of this task of societal importance.

\section{Related Work} \label{sec:prev}

The task of detecting cyberbullying content can be roughly divided into three categories. First, research with a focus on \emph{binary} classification, where it is only relevant if a message contains bullying or not. Second, more \emph{fine-grained} approaches where the task is to determine either the role of actors in a bullying scenario or the content type (i.e., different categories of bullying). Both binary and fine-grained approaches predominantly focus on text-based features. Lastly, \emph{meta-data} approaches that take more than just message content into account; these might include profile, network, or image information. Here, we will discuss efforts relevant to the task of cyberbullying classification within these three topics. We will predominantly focus on work conducted on openly available data, and those that report (positive) $F_1$-scores, to promote fair comparisons.\footnote{Unfortunately, numerous (recent) work on cyberbullying detection seems not to report such $F_1$-scores (in favor of accuracy), limit to criticized datasets with high baseline scores (such as the \textsc{caw} datasets) or do not show enough methodological rigor---some are therefore not included in this overview.} For an extensive literature review and a detailed comparison of different studies, see \cite{rosa2019automatic}.

\begin{table*}[t] \small

    \caption{Overview of datasets for cyberbullying detection. Lists the authors of the initial sets (Author), which work was conducted on it after publishing (Other), a reference name (Name), if it is publicly available and a link to its source (OS), which platform it was extracted from (Platform), the amount of reported cyberbullying instances (Pos), and that of non-cyberbullying (Neg), a reference to the work achieving the highest score on the data (Max) and what this (positive) $F_1$-score was ($F_1$). Please note that the instance numbers are as reported in the original work, and may have deviated through time (such as Twitter sets, and Formspring).}
    
  \begin{tabular*}{\textwidth}{@{\extracolsep{\fill}} lllllrrlr}
    \hline\noalign{\smallskip}
    Author         & Other                                   & Name              & OS                                                          &  Platform   & Pos & Neg & Max & $F_1$ \\
    \noalign{\smallskip}\hline\noalign{\smallskip}
    \cite{Yin2009} & \cite{Nahar2013}                        & \textsc{caw\_kon} & \href{http://caw2.barcelonamedia.org/}{v}                   & Kongregate & $42$    & $4802$  & \cite{Nahar2013}         & $.920$ \\
    \cite{Yin2009} & \cite{Nahar2013}                        & \textsc{caw\_sls} & \href{http://caw2.barcelonamedia.org/}{v}                   & Slashdot   & $60$    & $4303$  & \cite{Nahar2013}         & $.920$ \\
    \cite{Yin2009} & \cite{Nahar2013}                        & \textsc{caw\_msp} & \href{http://caw2.barcelonamedia.org/}{v}                   & Myspace    & $65$    & $1946$  & \cite{Nahar2013}         & $.920$ \\
    \cite{Reynolds2011} & \cite{Kontostathis2013,Squic2015,rosa2018using,rosa2018deeper}  & \textsc{kon\_frm} & \href{http://www.chatcoder.com/DataDownload}{v}             & Formspring & $369$   & $3915$  & \cite{rosa2018deeper}         & $.848$ \\
    \cite{Dinakar2011} & \cite{}                             & \textsc{din\_ytb} & x                                                           & YouTube    & $2277$  & $4500$  & \cite{Dinakar2011}       & $-$ \\
    \cite{Bayzick2011} & \cite{Zhao2016a,Squic2015}          & \textsc{bay\_msp} & \href{http://www.chatcoder.com/DataDownload}{v}             & Myspace    & $415$   & $1647$  & \cite{Zhao2016a}         & $.776$ \\
    \cite{Xu2012} & \cite{Zhao2016}                          & \textsc{xu\_trec} & \href{https://research.cs.wisc.edu/bullying/data.html}{v}  & Twitter    & $684$   & $1762$  & \cite{Zhao2016}          & $.780$ \\
    \cite{Dadvar2014} & \cite{}                              & \textsc{ddv\_msp} & x                                                           & Myspace    & $311$   & $8938$  & \cite{Dadvar2014}        & $.350$ \\
    \cite{Dadvar2014} & \cite{}                              & \textsc{ddv\_ytb} & x                                                           & YouTube    & $449$   & $4177$  & \cite{Dadvar2014}        & $.640$ \\
    \cite{Bret2014} & \cite{}                                & \textsc{brt\_twi} & \href{http://www.ub-web.de/research/}{v}                    & Twitter    & $220$   & $5162$  & \cite{Bret2014}          & $.726$ \\
    \cite{Bret2014} & \cite{}                                & \textsc{brt\_tw2} & \href{http://www.ub-web.de/research/}{v}                    & Twitter    & $194$   & $2599$  & \cite{Bret2014}          & $.719$ \\
    \cite{VanHee2015} & \cite{}                              & \textsc{ami\_ask} & \href{https://osf.io/rgqw8/}{v}                                                  & Ask.fm     & $3787$  & $86419$ & \cite{VanHee2015}        & $.465$ \\
    \cite{Hosseinmardi2015a} & \cite{cheng2019hierarchical}                       & \textsc{hos\_ins} & \href{https://sites.google.com/site/cucybersafety/home/
cyberbullying-detection-project/dataset}{v}                                                           & Instagram  & $567$   & $1387$  & \cite{cheng2019hierarchical} & $.783$ \\
    \cite{Sui2015} & \cite{Zhao2016a}                        & \textsc{sui\_twi} & \href{https://research.cs.wisc.edu/bullying/data.html.}{v}  & Twitter    & $2102$  & $5219$  & \cite{Zhao2016a}         & $.719$ \\
    \noalign{\smallskip}\hline
  \end{tabular*}
  
    \label{tab:dat}
\end{table*}

\subsection{Binary Classification}

One of the first traceable suggestions for applying text mining specifically to the task of cyberbullying detection is made by \cite{Kontostathis2010}, who note that \cite{Yin2009} previously tried to classify online harassment on the \textsc{caw} 2.0 dataset.\footnote{Data has been made available at \url{caw2.barcelonamedia.org}.} In the latter research, Yin et al.\ already state that the ratio of documents with harassing content to typical documents is challengingly small. Moreover, they foresee several other critical issues with regards to the task: a lack of positive instances will make detecting characteristic features a difficult task, and human labeling of such a dataset might have to face issues of ambiguity and sarcasm that are hard to assess when messages are taken out of conversation context. Even with very sparse datasets (with less than $1\%$ positive class instances), the harassment classifier outperforms the random baseline using tf$\cdot{}$idf, pronoun, curse word, and post similarity features. 

Following up \cite{Yin2009}, \cite{Reynolds2011} note that the \textsc{caw} 2.0 dataset is generally unfit for cyberbullying classification: in addition to lacking bullying labels (it only provides harassment labels), the conversations are predominantly between adults. Their work, along with \cite{Bayzick2011}, is a first effort to create datasets for cyberbullying classification through scraping the question-answering website Formspring.me, as well as Myspace.\footnote{Data has been made available at \url{www.chatcoder.com/DataDownload}.} In contrast with similar research, they aim to use textual features while deliberately avoiding Bag-of-Words (BoW) features. Through a curse word dictionary and custom severity annotations, they construct several metrics for features related to these ``bad'' words. In their more recent paper, \cite{Kontostathis2013} redid analyses on the \textsc{kon\_frm} set, primarily focusing on the contribution curse words have in the classification of bullying messages. By forming queries from curse word dictionaries, they show that there is no one combination which retrieves all. Moreover, using Essential Dimensions of Latent Semantic Indexing, they show potential for extracting messages containing harmful content, favoring high precision.

More recent efforts include \cite{Bret2014}, who combined word normalization, Named Entity Recognition to detect person-specific references, and multiple curse word dictionaries \cite{noswear,broadc,Millwood2000} in a rule-based pattern classifier, scoring well on Twitter data.\footnote{Data has been made available at \url{www.ub-web.de/research}.} Our own work \cite{VanHee2015}, where we collected a large dataset with posts from Ask.fm, used standard BoW features as a first test. Later, these were extended in \cite{vanhee2018} with term lists, subjectivity lexicons, and topic model features. Recently popularized techniques of word embeddings and neural networks have been applied by \cite{Zhao2016,Zhao2016a} on \textsc{xu\_trec}, \textsc{nay\_msp} and \textsc{sui\_twi}, both resulting in the highest performance for those sets. Convolutional Neural Networks (CNNs) on phonetic features were applied by \cite{zhang2016cyberbullying} and \cite{rosa2018deeper} investigate among others the same architecture on textual features in combination with Long Short Term Memory Networks (LSTMs). Both \cite{rosa2018deeper} and that of \cite{agrawal2018deep} investigate the C-LSTM \cite{zhou2015c}, the latter includes Synthetic Minority Over-sampling Technique (SMOTE). However, as we will show in the current research, both of these works suffer from reproducibility issues. Finally, fuzzified vectors of top-$k$ word lists for each class were used to conduct membership likelihood-based classification by \cite{rosa2018using} on \textsc{kon\_frm}, boosting recall over previously used methods.

\subsection{Fine-Grained Classification}

 The common denominator of the previously discussed research  was a focus on detecting single messages with evidence of cyberbullying per instance. The work of \cite{Xu2012} proposes a more fine-grained approach by looking at \emph{bullying traces}; i.e., the responses to a bullying incident. Their research is split up in a set of tasks on keyword-retrieved (\emph{bully}) Twitter data:\footnote{Data has been made available at \url{research.cs.wisc.edu/bullying/data.html}.} (1) a text classification task where solely relying on uni+bigram features yielded the best result, (2) a role labeling task, where semantic role labeling was then used to distinguish person-mention roles, (3) the incorporation of sentiment in the sentiment analysis task (3) to determine teasing, where despite high accuracy, $48\%$ of the positive instances were misclassified. Finally, (4) a latent topic modeling task, applying Latent Dirichlet Allocation to their corpus to note that some of the generated topics were relevant to bullying. Lastly, in our work, we demonstrated the difficulty of fine-grained approaches with simple BoW and sentiment features, especially detecting types of cyberbullying \cite{VanHee2015,VanHee2015guide}.

\subsection{Meta-data Features} \label{par:meta} A notable, yet less popular aspect of this task is the utilization of a graph for visualizing potential bullies and their connections. This method was first adopted by \cite{Nahar2013}, who use this information in combination with a classifier trained on LDA and weighted tf$\cdot$idf features to detect bullies and victims on the \textsc{caw\_*} datasets. Work that more concretely implements techniques from graph theory is that of \cite{Squic2015}, who used a wide range of features: network features to measure popularity (e.g., degree centrality, closeness centrality), content-based features, (length, sentiment, offensive words, second-person pronouns), and incorporated age, gender, and number of comments. They achieved the highest performance on \textsc{kon\_frm} and \textsc{bay\_msp}. 

Work by \cite{Hosseinmardi2015a} focuses on Instagram posts and incorporates platform-specific features retrieved from images and its network. They are the first adhere to the literature more closely and define cyberagression \cite{Kowalski2012b} separately from cyberbullying, in that these are single negative posts rather than the repeated character of cyberbullying. They also show that certain LIWC (Linguistic Inquiry and Word Count) categories, such as death, appearance, religion, and sexuality, give a good indication of cyberbullying. While BoW features perform best, meta-data features (such as user properties and image content) in combination with textual features from the top 15 comments achieve a similar score. Cyberagression seems to be slightly easier to classify.

\section{Task Evaluation Importance and Hypotheses} \label{sec:oos}

The domain of cyberbullying detection is in its early stages, as can be seen in Table~\ref{tab:dat}. Most datasets are quite small, and only a few have seen repeated experiments. Given the substantial societal importance of improving the methods developed so far, pinpointing shortcomings in the current state of research should assist in creating a robust framework under which to conduct future experiments---particularly concerning evaluating (domain) generalization of the classifiers. The latter of which, to our knowledge, none of the current research seems involved with. This is therefore the main focus of our work. In this section, we define three motivations for assessing this.

\subsection{Data Scarcity} \label{subs:scarcity}

Considering the complexity of the social dynamics underlying the target of classification, and the costly collection and annotation of training data, the issue of data scarcity can mostly be explained with respect to the aforementioned restrictions on data access: while on a small number of platforms most data is accessible without any internal access (commonly as a result of optional user anonymity), it can be assumed that a signifcant part of actual bullying takes place `behind closed doors'. To uncover this, one would require access to all known information within a social network (such as friends, connections, and private messages, including all meta-data). As this is unrealistic in practice, researchers rely on the small subset of publicly accessible data (predominantly text) streams. Consequently, most of the datasets used for cyberbullying detection are small and exhibit an extreme skew between positive and negative messages (as can be seen in Table~\ref{tab:dsc}). It is unlikely that these small sets accurately capture the language-use on a given platform, and generalizable linguistic features of the bullying instances even less so. We therefore hypothesize that \textbf{1}) the samples are underpowered in terms of accurately representing the substantial language variation between platforms, both in normal language-use and bullying-specific language-use.

\subsection{Task Definition} \label{subs:task}

Furthermore, we argue that this scarcity introduces issues with adherence to the definition of the task of cyberbullying. The chances of capturing the underlying dynamics of \emph{cyberbullying} (as defined in the literature) are slim with the message-level (i.e., using single documents only) approaches that the majority of work in the field has used up until now. The users in the collected sources have to be rash enough to bully in the open, and particular (curse) word use that would explain the effectiveness of dictionary and BoW-based approaches in previous research. Hence, we also hypothesize that \textbf{2}) the positive instances are biased; only reflecting a limited dimension of bullying. A more realistic scenario---where characteristics such as repetitiveness and power imbalance are taken into consideration---would require looking at the interaction between persons, or even profile instances rather than single messages, which, as we argued, is not generally available. The work found in the meta-data category (Section \ref{par:meta}) supports this argument with improved results using this information.

This theory regarding the definition (or operationalization) of this task is shared by Rosa et al., who pose that ``\emph{the most representative studies on automatic cyberbullying detection, published from 2011 onward, have conducted isolated online aggression classification}'' \cite[p. 341]{rosa2019automatic}. We will mainly focus on the shared notion that this framing is limited to verbal aggression; however, our focus will empirically assess its overlap with data framed to solely contain online toxicity data (i.e., online / cyberagression) to find concrete evidence.

\subsection{Domain Influence}

Enriching previous work with data such as network structure, interaction statistics, profile information, and time-based analyses might provide fruitful sources for classification and a correct operationalization of the task. However, they are also domain-specific, as not all social media have such a rich interaction structure. Moreover, it is arguably naive to assume that social networks such as Facebook (for which in an ideal case, all aforementioned information sources are available) will stay a dominant platform of communication. Recently, younger age groups have turned towards more direct forms of communication such as WhatsApp, Snapchat, or media-focused forms such as Instagram \cite{Smith2018}. This move implies more private and less affluent environments in which data can be accessed (resulting  in even more scarcity), and that further development in the field requires a critical evaluation of the current use of the available features, and ways to improve cross-domain generalization overall. This work, therefore, does not disregard textual features; they would still need to be considered as the primary source of information, while paying particular attention to the issues mentioned here. We further try to contribute towards this goal and hypothesize that \textbf{3}) crowdsourcing bullying content potentially decreases the influence of domain-specific language-use, allows for richer representations, and alleviates data scarcity.

\section{Data} \label{sec:data}

\begin{table*} \small
        \caption{Corpus statistics for English and Dutch cyberbullying datasets, list number of positive (Pos, bullying) and negative (Neg, other) instances, Types (unique words), Tokens (total words), average number of tokens per message (Avg Tok/Msg), number of emojis and emoticons (Emotes), and swear word occurrence (Swears).\protect\footnotemark}
      \begin{tabular*}{\textwidth}{@{\extracolsep{\fill}} lrrrrrlrr}
      \hline\noalign{\smallskip}
       & Pos & Neg & Types & Tokens & \multicolumn{2}{c}{Avg Tok/Msg}  & Emotes & Swears                        \\ 
      \noalign{\smallskip}\hline\noalign{\smallskip}
        $D_{twB}$    & 236     & 5,237   & 11,650   & 78,649     & 14  & ($\sigma = 8$)    & 959     & 1,137   \\
        $D_{frm}$     & 1,024   & 11,711  & 20,818   & 383,968    & 30  & ($\sigma = 31$)   & 3,308   & 4,097   \\
        $D_{msp}$     & 426     & 1,633   & 13,586   & 828,583    & 402 & ($\sigma = 291$)  & 932     & 5,203   \\
        $D_{ytb}$     & 416     & 3,022   & 52,422   & 826,883    & 241 & ($\sigma = 252$)  & 3,664   & 1,1277  \\
        $D_{ask}$     & 4,951   & 95,159  & 61,640   & 1,156,345  & 12  & ($\sigma = 23$)   & 17,801  & 16,888  \\
        $D_{twX}$    & 281     & 4,703   & 18,754   & 87,582     & 18  & ($\sigma = 8$)    & 1,344   & 450     \\
        \noalign{\smallskip}\hline\noalign{\smallskip}
        $D_{tox}$     & 15,292  & 144,274 & 223,728  & 13,319,795 & 83  & ($\sigma = 125$)  & 11,813  & 36,118  \\ 
        \noalign{\smallskip}\hline\noalign{\smallskip}
        $D_{ask\_nl}$ & 8,055   & 66,328  & 67,135   & 814,970    & 10  & ($\sigma = 14$)   & 15,392  & 8,296   \\
        $D_{sim\_nl}$ & 2,343   & 2,546   & 7,039    & 62,340     & 12  & ($\sigma = 18$)   & 437     & 626     \\
        $D_{don\_nl}$ & 152     & 211     & 1,964    & 7,371      & 20  & ($\sigma = 24$)   & 32      & 79      \\
        \noalign{\smallskip}\hline
    \end{tabular*}
    \label{tab:dsc}
\end{table*}

\footnotetext{Emojis were detected with \url{https://github.com/NeelShah18/emot}. Swears were detected with reference lists: for English these were taken from \url{www.noswearing.com} and the Dutch were manually composed.}

For the current research, we distinguish a large variety of datasets. For those provided through the AMiCA (Automatic  Monitoring in Cyberspace Applications)\footnote{\url{www.amicaproject.be}} project, the \emph{Ask.fm} corpus is partially available open-source,\footnote{\url{https://osf.io/rgqw8/}} and the \emph{Crowdsourced} corpus will be made available upon request. All other sources are publicly available datasets gathered from previous research\footnote{These were collected as complete as possible. Twitter, in particular, has low recall; only an approximate of 60\% of the tweets were retrieved. Such numbers are expected given the classification problem; people tend to remove harassing messages as was shown before by \cite{Xu2012}.} as discussed in Section~\ref{sec:prev}. Corpus statistics of all data discussed below can be found in Table~\ref{tab:dsc}. The sets' abbreviations, language (\textsc{en} for English, \textsc{nl} for Dutch), and brief collection characteristics can be found below.

\subsection{AMiCA}

\paragraph{Ask.fm} ($D_{ask}$, $D_{ask\_nl}$, \textsc{en}, \textsc{nl}) were collected from the eponymous social network by \cite{VanHee2015}. Ask.fm is a question answering-style network where users interact by (frequently anonymously) asking questions on other profiles, and answering questions on theirs. As such, a third party cannot react to these question-answer pairs directly. The anonymity and restrictive interactions make for a high amount of potential cyberbullying. Profiles were retrieved through profile seed list, used as a starting point for traversing to other profiles and collecting all existing question-answer pairs for those profiles---these are predominantly Dutch and English. Each message was annotated with fine-grained labels (further details can be found in \cite{VanHee2015guide}); however, for the current experiments these were binarized, with any form of bullying being labeled positive.
    
\paragraph{Donated} ($D_{don\_nl}$, \textsc{nl}) contains instances of (Dutch) cyberbullying from a mixture of platforms such as Skype, Facebook, and Ask.fm. The set is quite small; however, it contains several hate pages that are valuable collections of cyberbullying directed towards one person. The data was donated for use in the AMiCA project by previously bullied teens, thus forming a reliable source of gold standard, real-life data.
    
\paragraph{Crowdsourced} ($D_{sim\_nl}$, \textsc{nl}) originates from a crowdsourcing experiment conducted by \cite{Broeck2014}, wherein 200 adolescents aged 14 to 18 partook in a role-playing experiment on an isolated SocialEngine\footnote{\url{www.socialengine.com}} social network. Here, each respondent was given the account of a fictitious person and put in one of four roles in a group of six: a bully, a victim, two bystander-assistants, and two bystander-defenders. They were asked to read---and identify with---a character description and respond to an artificially generated initial post attributed to one of the group members. All were confronted with two initial posts containing either  low- or high-perceived severity of cyberbullying.

\subsection{Related Work}

\paragraph{Formspring} ($D_{frm}$, \textsc{en}) is taken from the research by \cite{Reynolds2011} and is composed of posts from Formspring.me, a question-answering platform similar to Ask.fm. As Formspring is mostly used by teenagers and young adults, and also provides the option to interact anonymously, it is notorious for hosting large amounts of bullying content \cite{Binns2013}. The data was annotated through Mechanical Turk, providing a single label by majority vote for a question-answer pair. For our experiments, the question and answer pairs were merged into one document instance.

\paragraph{Myspace} ($D_{msp}$, \textsc{en}) was collected by \cite{Bayzick2011}. As this was set up as an information retrieval task, the posts are labeled in batches of ten posts, and thus a single label applies to the entire batch (i.e., does it include cyberbullying). These were merged per batch as one instance and labeled accordingly. Due to this batching, the average tokens per instance are much higher than any of the other corpora.
    
\paragraph{Twitter} ($D_{twB}$, \textsc{en}) by \cite{Bret2014} was collected from the stream between 20-10-2012 and 30-12-2012, and was labeled based on a majority vote between three annotators. Excluding re-tweets, the main dataset consists of 220 positive and 5162 negative examples, which adheres to the general expected occurrence rate of 4\%. Their comparably-sized test set, consisting of 194 positive and 2699 negative examples, was collected by adding a filter to the stream for messages to contain any of the words \texttt{school}, \texttt{class}, \texttt{college}, and \texttt{campus}. These sets are merged for the current experiments.
    
\paragraph{Twitter II} ($D_{twX}$, \textsc{en}) from \cite{Xu2012} focussed on \emph{bullying traces}, and was thus retrieved by keywords (\texttt{bully}, \texttt{bullying}), which if left unmasked generates a strong bias when utilized for classification purposes (both by word use as well as being a mix of toxicity and victims). It does, however, allow for demonstrating the ability to detect bullying-associated topics, and (indirect) reports of bullying.

\subsection{Experiment-specific}

\paragraph{Ask.fm Context} ($C_{ask}$, $C_{ask}\_nl$, \textsc{en}, \textsc{nl}) --- the Ask.fm corpus was collected on profile level, but prior experiments have focused on single message instances \cite{vanhee2018}. Here, we aggregate all messages for a single profile, which is then labeled as positive when as few as a single bullying instance occurs on the profile. This aggregation shifts the task of cyberbullying message detection to victim detection on profile level, allowing for more access to context and profile-level severity (such as repeated harassment), and makes for a more balanced set (1,763 positive and 6,245 negative instances). 

\paragraph{Formspring Context} ($C_{frm}$, \textsc{en}) --- similar to the Ask.fm corpus, was collected on profile level \cite{Reynolds2011}. However, the set only includes 49 profiles, some of which only include a single message. Grouping on full profile level would result in very few instances; thus, we opted for creating small `context' in batches of five (of the same profile). Similar to the Ask.fm approach, if one of these messages contains bullying, it is labeled positive, balancing the dataset (565 positive and 756 negative instances). 

\paragraph{Toxicity} ($D_{tox}$, \textsc{en}) from Kaggle\footnote{\url{https://www.kaggle.com/c/jigsaw-toxic-comment-classification-challenge}} is a Toxic Comment Classification dataset created by Conversation AI\footnote{\url{https://conversationai.github.io/}} \cite{Thain2017} which offers over 300k messages from Wikipedia comments with Crowdflower-annotated labels for toxicity (including subtypes). Noteworthy is how \emph{disjoint} both the task and the platform are from the rest of the corpora used in this research. While toxicity shares many properties with bullying, the focus here is on single instances of insults directed to likely unknown people (to the harasser). Given Wikipedia as a source, the article and moderation focussed comments make it topically quite different from what one would expect on social media---the fundamental overlap being curse words, which is only one of many dimensions to be captured to detect cyberbullying (as opposed to toxicity). 

\subsection{Preprocessing} \label{subs:proc}

All texts were tokenized using spaCy \cite{spacy2}.\footnote{\url{https://spacy.io} (\texttt{v2.0.5})} No preprocessing was conducted for the corpus statistics in Table~\ref{tab:dsc}. All models (Section~\ref{sec:exp}) applied lowercasing and special character removal only; other preprocessing decreased performance (see Table~\ref{tab:archs}).

\begin{figure*}
    \begin{minipage}{.50\textwidth}
    \begin{tikzpicture}
  \begin{axis}[enlargelimits=false,
  			     			   colorbar,
                                                            tick pos=left,
               tick align=outside,
               point meta min=0,
               colormap={whiteblue}{color=(white) color=(black)},
               try min ticks=8,
               xticklabels={,$D_{twiB}$,$D_{frm}$,$D_{msp}$,$D_{ytb}$,$D_{ask}$,$D_{twiX}$,$D_{frmC}$,$D_{askC}$,$D_{tox}$},
               yticklabels={,$D_{twiB}$,$D_{frm}$,$D_{msp}$,$D_{ytb}$,$D_{ask}$,$D_{twiX}$,$D_{frmC}$,$D_{askC}$,$D_{tox}$},
                              xticklabel style = {rotate=90},
               			   			   ]
    \addplot [matrix plot,point meta=explicit]
      coordinates {
        (0,0) [0.16] (1,0) [0.29] (2,0) [0.22] (3,0) [0.31] (4,0) [0.30] (5,0) [0.11] (6,0) [0.28] (7,0) [0.25] (8,0) [0.08]

        (0,1) [0.05] (1,1) [0.18] (2,1) [0.18] (3,1) [0.25] (4,1) [0.24] (5,1) [0.04] (6,1) [0.20] (7,1) [0.13] (8,1) [0.15]

        (0,2) [0.03] (1,2) [0.08] (2,2) [0.32] (3,2) [0.15] (4,2) [0.12] (5,2) [0.02] (6,2) [0.08] (7,2) [0.06] (8,2) [0.20]
        
        (0,3) [0.03] (1,3) [0.08] (2,3) [0.14] (3,3) [0.18] (4,3) [0.14] (5,3) [0.03] (6,3) [0.08] (7,3) [0.07] (8,3) [0.22]

        (0,4) [0.02] (1,4) [0.07] (2,4) [0.11] (3,4) [0.14] (4,4) [0.18] (5,4) [0.02] (6,4) [0.07] (7,4) [0.08] (8,4) [0.17]
        
        (0,5) [0.11] (1,5) [0.20] (2,5) [0.16] (3,5) [0.24] (4,5) [0.21] (5,5) [0.15] (6,5) [0.19] (7,5) [0.18] (8,5) [0.07]
        
        (0,6) [0.05] (1,6) [0.19] (2,6) [0.19] (3,6) [0.25] (4,6) [0.24] (5,6) [0.04] (6,6) [0.18] (7,6) [0.13] (8,6) [0.15]
        
        (0,7) [0.05] (1,7) [0.15] (2,7) [0.15] (3,7) [0.23] (4,7) [0.35] (5,7) [0.04] (6,7) [0.15] (7,7) [0.17] (8,7) [0.11]
        
        (0,8) [0.00] (1,8) [0.01] (2,8) [0.03] (3,8) [0.03] (4,8) [0.02] (5,8) [0.00] (6,8) [0.01] (7,8) [0.01] (8,8) [0.19]
      };
      \node[color=black] at (axis cs:0,0) {0.16};
      \node[color=white] at (axis cs:1,0) {0.29};
      \node[color=white] at (axis cs:2,0) {0.22};
      \node[color=white] at (axis cs:3,0) {0.31};
      \node[color=white] at (axis cs:4,0) {0.30};
      \node[color=black] at (axis cs:5,0) {0.11};
      \node[color=white] at (axis cs:6,0) {0.28};
      \node[color=white] at (axis cs:7,0) {0.25};
      \node[color=black] at (axis cs:8,0) {0.08};
      
      \node[color=black] at (axis cs:0,1) {0.05};
      \node[color=black] at (axis cs:1,1) {0.18};
      \node[color=black] at (axis cs:2,1) {0.18};
      \node[color=white] at (axis cs:3,1) {0.25};
      \node[color=white] at (axis cs:4,1) {0.24};
      \node[color=black] at (axis cs:5,1) {0.04};
      \node[color=white] at (axis cs:6,1) {0.20};
      \node[color=black] at (axis cs:7,1) {0.13};
      \node[color=black] at (axis cs:8,1) {0.15};
      
      \node[color=black] at (axis cs:0,2) {0.03};
      \node[color=black] at (axis cs:1,2) {0.08};
      \node[color=white] at (axis cs:2,2) {0.32};
      \node[color=black] at (axis cs:3,2) {0.15};
      \node[color=black] at (axis cs:4,2) {0.12};
      \node[color=black] at (axis cs:5,2) {0.02};
      \node[color=black] at (axis cs:6,2) {0.08};
      \node[color=black] at (axis cs:7,2) {0.06};
      \node[color=white] at (axis cs:8,2) {0.20};
      
      \node[color=black] at (axis cs:0,3) {0.03};
      \node[color=black] at (axis cs:1,3) {0.08};
      \node[color=black] at (axis cs:2,3) {0.14};
      \node[color=black] at (axis cs:3,3) {0.18};
      \node[color=black] at (axis cs:4,3) {0.14};
      \node[color=black] at (axis cs:5,3) {0.03};
      \node[color=black] at (axis cs:6,3) {0.08};
      \node[color=black] at (axis cs:7,3) {0.07};
      \node[color=white] at (axis cs:8,3) {0.22};
      
      \node[color=black] at (axis cs:0,4) {0.02};
      \node[color=black] at (axis cs:1,4) {0.07};
      \node[color=black] at (axis cs:2,4) {0.11};
      \node[color=black] at (axis cs:3,4) {0.14};
      \node[color=black] at (axis cs:4,4) {0.18};
      \node[color=black] at (axis cs:5,4) {0.02};
      \node[color=black] at (axis cs:6,4) {0.07};
      \node[color=black] at (axis cs:7,4) {0.08};
      \node[color=black] at (axis cs:8,4) {0.17};
      
      \node[color=black] at (axis cs:0,5) {0.11};
      \node[color=white] at (axis cs:1,5) {0.20};
      \node[color=black] at (axis cs:2,5) {0.16};
      \node[color=white] at (axis cs:3,5) {0.24};
      \node[color=white] at (axis cs:4,5) {0.21};
      \node[color=black] at (axis cs:5,5) {0.15};
      \node[color=black] at (axis cs:6,5) {0.19};
      \node[color=black] at (axis cs:7,5) {0.18};
      \node[color=black] at (axis cs:8,5) {0.07};
      
      \node[color=black] at (axis cs:0,6) {0.05};
      \node[color=black] at (axis cs:1,6) {0.19};
      \node[color=black] at (axis cs:2,6) {0.19};
      \node[color=white] at (axis cs:3,6) {0.25};
      \node[color=white] at (axis cs:4,6) {0.24};
      \node[color=black] at (axis cs:5,6) {0.04};
      \node[color=black] at (axis cs:6,6) {0.18};
      \node[color=black] at (axis cs:7,6) {0.13};
      \node[color=black] at (axis cs:8,6) {0.15};
      
      \node[color=black] at (axis cs:0,7) {0.05};
      \node[color=black] at (axis cs:1,7) {0.15};
      \node[color=black] at (axis cs:2,7) {0.15};
      \node[color=white] at (axis cs:3,7) {0.23};
      \node[color=white] at (axis cs:4,7) {0.35};
      \node[color=black] at (axis cs:5,7) {0.04};
      \node[color=black] at (axis cs:6,7) {0.15};
      \node[color=black] at (axis cs:7,7) {0.17};
      \node[color=black] at (axis cs:8,7) {0.11};
      
      \node[color=black] at (axis cs:0,8) {0.00};
      \node[color=black] at (axis cs:1,8) {0.01};
      \node[color=black] at (axis cs:2,8) {0.03};
      \node[color=black] at (axis cs:3,8) {0.03};
      \node[color=black] at (axis cs:4,8) {0.02};
      \node[color=black] at (axis cs:5,8) {0.00};
      \node[color=black] at (axis cs:6,8) {0.01};
      \node[color=black] at (axis cs:7,8) {0.01};
      \node[color=black] at (axis cs:8,8) {0.19};

      \clip (current bounding box.south west) rectangle (current bounding box.north east);
  \end{axis}
  
\end{tikzpicture}
    \caption{Jaccard similarity between training sets ($y$-axis) and test sets ($x$-axis).}
    \label{fig:jac}
    \end{minipage} \hfill
    \begin{minipage}{.42\textwidth} \small 
      \caption{SVM \texttt{baseline} and NBSVM grid values used in hyper-parameter search. \smallskip}
      \centering
      \resizebox{1.0\textwidth}{!}{
      \begin{tabular*}{\textwidth}{llp{3cm}}
      \hline\noalign{\smallskip}
      Part & Params & Values   \\ 
      \noalign{\smallskip}\hline\noalign{\smallskip}
      BoW  & range & $(1, 1)$, $(1, 2)$, $(1, 3)$     \\
                    & level          & words                       \\
      \noalign{\smallskip}\hline\noalign{\smallskip}
      SVM  & weight $y$   & default, balanced                    \\
                               & loss           & hinge, square hinge \\
                               & $C$            & $1\mathrm{e}{-3}$, $1\mathrm{e}{-2}$, $\ldots$, $1\mathrm{e}{2}$, $1\mathrm{e}{3}$ \\
      \noalign{\smallskip}\hline
      \strut \\
      \strut \\
      \strut \\
      \strut \\
    \end{tabular*}}
    \label{tab:hyp}
    \end{minipage}
\end{figure*}

\subsection{Descriptive Analysis} \label{subs:sum}

Both Table~\ref{tab:dsc} and Figure~\ref{fig:jac} illustrate stark differences; not only across domains but more importantly, between in-domain training and test sets. Most do not exceed a Jaccard similarity coefficient over 0.20 (Figure~\ref{fig:jac}), implying a large part of their vocabularies do not overlap. This contrast is not necessarily problematic for classification; however, it does hamper learning a general representation for the negative class. It also clearly illustrates how even more disjoint $D_{twX}$ (collected by trace queries) and $D_{tox}$ are from the rest of the corpora and splits. Finally, the descriptives (Table~\ref{tab:dsc}) further show significant differences in size, message length, class balance, and type/token ratios (i.e., writing level). In conclusion, it can be assumed that the language-use in both positive as negative instances will vary significantly, and that it will be challenging to model in-domain, and generalize out-of-domain.

\section{Experimental Setup} \label{sec:exp}

We attempt to address the hypotheses posited in Section~\ref{sec:oos} and propose five main experiments. Experiments I and III deal with the problem of generalizability, whereas Experiment II and V will both propose a solution for restricted data collection. Experiment IV will reproduce a selection of the current state-of-the-art models for cyberbullying detection and subject them to our cross-domain evaluation, to be compared against our baselines.

\subsection{Experiment I: Cross-Domain Evaluation} \label{subs:crit}

In this experiment, we introduce the cross-domain evaluation framework, which will be extended in all other experiments. For this, we initially perform a many-to-many evaluation of a given model (baseline or otherwise) trained individually on all available data sources, split in train and test. In later experiments, we extend this with a one-to-many evaluation. This setup implies that (i) we fit our model on some given corpus' training portion and evaluate prediction performance on all available corpora their test portions (many-to-many) individually. Furthermore, we (ii) fit on all corpora their train portions combined, and evaluate on all their test portions individually (one-to-many). In sum, we report on `small' models trained on each corpus individually, as well as a `large' one trained on them combined, for each test set individually.

For every experiment, hyper-parameter tuning was conducted through an exhaustive grid search, using nested cross-validation (with ten inner and three outer folds) on the training set to find the optimal combination of the given parameters. Any model selection steps were based on the evaluation of the outer folds. The best performing model was then refitted on the full training set (90\% of the data) and applied to the test set (10\%). All splits (also during cross-validation) were made in a stratified fashion, keeping the label distributions across splits similar to the whole set. Henceforth, all experiments in this section can be assumed to follow this setup. 

The many-to-many evaluation framework intends to test Hypothesis \textbf{1} (Section \ref{subs:scarcity}), relating to language variation and cross-domain performance of cyberbullying detection. To facilitate this, we employ an initial \texttt{baseline} model: Scikit-learn's \cite{scikit-learn2011} Linear Support Vector Machine (SVM) \cite{Cortes1995,Fan2008} implementation trained on binary BoW features, tuned using the grid shown in Table~\ref{tab:hyp}, based on \cite{vanhee2018}. Given its use in previous research, it should form a strong candidate against which to compare. To ascertain out-of-domain performance compared to this baseline, we report test score averages across all test splits, excluding the set the model was trained on (in-domain).

Consequently, we add an evaluation criterion to that of related work: a model should both perform overall best in-domain and achieve the highest out-of-domain performance on average to classify as a robust method. It should be noted that the selected corpora for this work are not all optimally representative for the task. The tests in our experiments should, therefore, be seen as an initial proposal to improve the task evaluation.

\subsection{Experiment II: Gauging Domain Influence}

In an attempt to overcome domain restrictions on language-use, and to further solidify our tests regarding Hypothesis \textbf{1}, we aim to improve the performance of our baseline models through changing our representations in three distinct ways: i) merging all available training sets (as to simulate a large, diverse corpus), ii) by aggregating instances on user-level, and iii) using state-of-the-art language representations over simple BoW features in all settings. We define these experiments as such:

\paragraph{Volume and Variety} Some corpora used for training are relatively small, and can thus be assumed insufficient to represent held-out data (such as the test sets). One could argue that this can be partially mitigated through simply collecting more data or training on multiple domains. To simulate such a scenario, we merge all available cyberbullying-related training splits (creating $D_{all}$), which then corresponds to the one-to-many setting of the evaluation framework. The hope is that corpora similar in size or content (the Twitter sets, Ask.fm and Formspring, YouTube and Myspace) would benefit from having more (related) data available. Additionally, training a large model on its entirety facilitates a catch-all setting for assessing the average cross-domain performance of the full task (i.e. across all test sets when trained on all available corpora). This particular evaluation will be used in Experiment IV (replication) for model comparison.

\paragraph{Context Change} Practically all corpora, save for MySpace and YouTube, have annotations based on short sentences, which is particularly noticeable in Table~\ref{tab:dsc}. This one-shot (i.e., based on a single message) method of classifying cyberbullying provides minimal content (and context) to work with. It does therefore not follow the definition of cyberbullying---as previously discussed in Section~\ref{subs:task}. As a preliminary simulation\footnote{Preferably, one would want to collect data on profile level by design. The corpora available were not specifically collected this way, making our set-up an approximation of such a setting.} of adding (richer) context, we merge the profiles of $D_{ask}$ and (batches of) $D_{frm}$ into single context instances (creating $C_{ask}$ and $C_{frm}$, see Section~\ref{sec:data}). This allows us to compare models trained larger contexts directly to that of single messages, and evaluate how context restrictions affect performance on the task in general, as well as cross-domain.

\paragraph{Improving Representations}

Pre-trained word embeddings as language representation have been demonstrated to yield significant performance gains for a multitude of NLP-related tasks \cite{collobert2011natural}. Given the general lack of training data---including negative instances for many corpora---word features (and weightings) trained on the available data tend to be a poor reflection of the language-use on the platform itself, let alone other social media platforms. Therefore, pre-trained semantic representations provide features that in theory, should perform better in cross-domain settings. We consider two off-the-shelf embedding models per language that are suitable for the task at hand: for English, averaged 200-dimensional GloVe \cite{pennington2014glove} vectors trained on Twitter\footnote{\url{https://nlp.stanford.edu/projects/glove/} (\texttt{v1.2})}, and DistilBERT \cite{sanh2019distilbert} sentence embeddings\footnote{\url{https://github.com/huggingface/transformers} (\texttt{1d646ba})} \cite{devlin2018bert}. For Dutch, \texttt{fastText} embeddings \cite{bojanowski2017enriching} trained on Wikipedia\footnote{\url{https://github.com/facebookresearch/fastText/blob/master/pretrained-vectors.md} (\texttt{2665eac})} and \texttt{word2vec} \cite{mikolov2013efficient,mikolov2013distributed} embeddings\footnote{\url{https://github.com/clips/dutchembeddings} (\texttt{1e3d528})} \cite{tulkens2016evaluating} trained on the COrpora from the Web (COW) corpus \cite{schafer2012building} embeddings. The GLoVe, \texttt{fastText}, and \texttt{word2vec} embeddings were processed using Gensim\footnote{\url{https://radimrehurek.com/gensim/index.html} (\texttt{v3.4})} \cite{rehurek_lrec}. 

As an additional baseline for this section, we include the Naive Bayes Support Vector Machine (NBSVM) from \cite{wang2012baselines}, which should offer competitive performance on text classification tasks.\footnote{The implementations for these models can be found in our repository.} This model also served as a baseline for the Kaggle challenge related to $D_{tox}$.\footnote{\url{https://kaggle.com/jhoward/nb-svm-strong-linear-baseline/notebook}} NBSVM uses tf$\cdot$idf-weighted uni and bi-gram features as input, with a minimum document frequency of 3, and corpus prevalence of 90\%. The idf values are smoothed and tf scaled sublinearly ($1 + \log($tf$)$). These are then weighted by their log-count ratios derived from Multinomial Naive Bayes. 

Tuning of both embeddings and NB representation classifiers is done using the same grid as Table~\ref{tab:hyp}, however replacing $C$ with $[1, 2, 3, 4, 5, 10, 25, 50, 100, 200, 500]$. Lastly, we opted for Logistic Regression (LR), primarily as this was used in the NBSVM implementation mentioned above, as well as \texttt{fastText}. Moreover, we found SVM using our grid to perform marginally worse using these features. The embeddings were not fine-tuned for the task. While this could potentially increase performance, it complicates direct comparison to our baselines---we leave this for Experiment IV.

\subsection{Experiment III: Aggression Overlap} 

In previous research using fine-grained labels for cyberbullying classification (e.g., \cite{vanhee2018}) it was observed that cyberbullying classifiers achieve the lowest error rates on blatant cases of aggression (cursing, sexual talk, and threats), an idea that was further adopted by \cite{rosa2019automatic}. To empirically test Hypothesis \textbf{2} (see Section~\ref{subs:task})---related to the bias present in the available positive instances---we adapt the idea of running a profanity baseline from this previous work. However, rather than relying on look-up lists containing profane words, we expand this idea by training a separate classifier on toxicity detection ($D_{tox}$) and seeing how well this performs on our bullying corpora (and vice-versa). For the corpora with fine-grained labels, we can further inspect and compare the bullying classes captured by this model. 

We argue that high test set performance overlap of a toxicity detection model with models trained on cyberbullying detection gives strong evidence of nuanced aspects of cyberbullying not being captured by such models. Notably, in line with \cite{rosa2019automatic}, that the current operationalization does not significantly differ from the detection of online aggression (or toxicity)---and therefore does not capture actual cyberbullying. Given enough evidence, both issues should be considered as crucial points of improvement for the further development of classifiers in this domain.

\subsection{Experiment IV: Replicating State-of-the-Art}

For this experiment, we include two architectures that achieved state-of-the-art results on cyberbullying detection. As a reference neural network model for language-based tasks, we used a Bidirectional \cite{schuster1997bidirectional,baldi1999exploiting} Long Short-Term Memory network \cite{hochreiter1997long,gers2002learning} (BiLSTM), partly reproducing the architecture from \cite{agrawal2018deep}. We then attempt to reproduce the Convolutional Neural Network (CNN) \cite{kim2014convolutional} used in both \cite{rosa2018deeper} and \cite{agrawal2018deep}, and the Convolutional LSTM (C-LSTM) \cite{zhou2015c} used in \cite{rosa2018deeper}. As \cite{rosa2018deeper} do not report essential implementation details for these models (batch size, learning rate, number of epochs), there is no reliable way to reproduce their work. We will, therefore, take \cite{agrawal2018deep} their implementation for the BiLSTM and CNN as the initial setup. Given that this work is available open-source, we run the exact architecture (including SMOTE) in our Experiment I and II evaluations. The architecture-specific details are as follows:

\paragraph{Reproduction} 

We initially adopt the basic implementation\footnote{\url{https://github.com/sweta20/Detecting-Cyberbullying-Across-SMPs/blob/master/DNNs.ipynb}}  by \cite{agrawal2018deep}: randomly initialized embeddings with a dimension of 50 (as the paper did not find significant effects of changing the dimension, nor initialization), run for 10 epochs with a batch size of 128, dropout probability of 0.25, and a learning rate of 0.01. Further architecture details can be found in our repository.\footnote{\url{https://github.com/cmry/amica/blob/master/neural.py}} We also run a variant with SMOTE on, and one from the provided notebooks directly.\footnote{Note this is for testing reproduction only, as it is not subjected to the same evaluation framework.} This and following neural models were run on an NVIDIA Titan X Pascal, using Keras \cite{chollet2015keras} with Tensorflow \cite{tensorflow2015whitepaper} as backend.

\paragraph{BiLSTM} For our own version of the BiLSTM, we minimally changed the architecture from \cite{agrawal2018deep}, only tuning using a grid on batch size $[32,$ $64,$ $128,$ $256]$, embedding size $[50,$ $100,$ $200,$ $300]$, and learning rate $[0.1,$ $0.01,$ $0.05,$ $0.001,$ $0.005]$. Rather than running for ten epochs, we use a validation split (10\% of the train set) and initiate early stopping when the validation loss does not go down after three epochs. Hence---and in contrast to earlier experiments---we do not run the neural models in 10-fold cross-validation, but a straightforward 2-fold train and test split where the latter is 10\%  of a given corpus. Again, we are predominantly interested in confirming statements made in earlier work; namely, that for this particular setting tuning of the parameters does not meaningfully affect performance.

\paragraph{CNN} 

We use the same experimental setup as for the BiLSTM. The implementations of \cite{agrawal2018deep,rosa2018deeper} use filter window sizes of 3, 4, and 5---max pooled at the end. Given that the same grid is used, the word embedding sizes are varied and weights trained (whereas \cite{rosa2018deeper} use 300-dimensional pre-trained embeddings). Therefore, for direct performance comparisons, \cite{agrawal2018deep} their results will be used as a reference. As CNN-based architectures for text classification are often also trained on character level, we include a model variant with this input as well. 

\paragraph{C-LSTM} For this architecture, we take an open-source text classification survey implementation.\footnote{\url{https://github.com/bicepjai/Deep-Survey-Text-Classification/}} This uses filter windows of [10, 20, 30, 40, 50], 64-dimensional LSTM cells and a final 128 dimensional dense layer. Please refer to our repository for additional implementation details---for this and previous architectures.

\subsection{Experiment V: Crowdsourced Data}

Following up on the proposed shortcomings of the currently available corpora in Hypotheses \textbf{1} and \textbf{2}, we propose the use of a crowdsourcing approach to data collection. In this experiment, we will repeat Experiment I and II with the best out-of-domain classifier from the above evaluations with three (Dutch\footnote{On account of the synthetic data being available in Dutch only. Experiment III was not repeated as there is no equivalent toxicity dataset available in this language.}) datasets: $D_{ask\_nl}$; the Dutch part of the Ask.fm dataset used before, $D_{sim\_nl}$; our synthetic, crowdsourced cyberbullying data, and lastly $D_{don\_nl}$; a small donated cyberbullying test set with messages from various platforms (full overview and description of these three can be found in Section~\ref{sec:data}). The only notable difference to our setup for this experiment is that we never use $D_{don\_nl}$ as training data. Therefore rather than $D_{all}$, the Ask.fm corpus is merged with the crowdsourced cyberbullying data to make up the $D_{comb}$ set. 

\section{Results and Discussion}

We will now cover results per experiment, and to what extent these provide support for the hypotheses posed in Section~\ref{sec:oos}. As most of these required backward evaluation (e.g., Experiment III was tested on sets from Experiment I), the results of Experiment I-III are compressed in Table~\ref{tab:base}.  Table~\ref{tab:archs} comprises the \emph{Improving Representations} part of Experiment II (under `word2vec' and `DistilBERT') along with the preprocessing results effect of our baselines. The results of Experiment V can be found in Table~\ref{tab:neural}. For brevity of reporting, the latter two only report on the in-domain scores, and feature the out-of-domain \emph{averages} for the $D_{all}$ models for comparison, and $D_{tox}$ averages in Table~\ref{tab:neural}.

\subsection{Experiment I}

\begin{table} \small
        \caption{Cross-corpora positive class $F_1$ scores for Experiment I (T1), II (T2), and III (T3). Models are fitted on the training proportion of the corpora row-wise, and tested column-wise. The out-of-domain average (Avg) excludes test performance of the parent training corpus. The best overall test score is noted in bold, the best out-of-domain performance in gray.} \label{tab:base}
        \begin{tabular*}{\textwidth}{@{\extracolsep{\fill}} lrrrrrrrrrr}
            \hline\noalign{\smallskip}
            Train         &                             \multicolumn{6}{c}{T1}                         & Avg         & \multicolumn{2}{c}{T2} & T3 \\
                          \cmidrule{2-7} \cmidrule{8-8} \cmidrule{9-10} \cmidrule{11-11}
                          & $D_{twB}$ & $D_{frm}$        & $D_{msp}$     & $D_{ytb}$ & $D_{ask}$ & $D_{twX}$ &         & $C_{frm}$ & $C_{ask}$ & $D_{tox}$  \\ 
            \noalign{\smallskip}\hline\noalign{\smallskip}
            $D_{twB}$    & .417          & .308          & .000          & .122      & .298      & .051      & .153      & .131      & .158      & .349 \\
            $D_{frm}$    & .423          & .454          & .042          & .379      & .418      & .041      & .321      & .682      & .259      & .465 \\
            $D_{msp}$    & .120          & .176          & {\bf.941}     & .324      & .168      & .043      & .197      & .364      & .185      & .185 \\
            $D_{ytb}$    & .074          & .160          & .375          & .365      & .138      & .000      & .183      & .338      & .197      & .140 \\
            $D_{ask}$    & .493          & .444          & .211          & \bf{.421} & \bf{.561} & .139      & .351      & .389      & .357      & .584 \\ 
            $D_{twX}$    & .049          & .131          & .184          & .175      & .077      & {\bf.508} & .205      & .496      & .325      & .082 \\ 
            \noalign{\smallskip}\hline\noalign{\smallskip}
            $D_{all}$    & \gc{\bf.524}  & \gc{\bf.473}  & \gc{\bf.941}  & \gc{.397} & \gc{.553} & \gc{.194} & \gc{\bf.557} & \gc{\bf.780} & \gc{.570} & \gc{.587} \\
            \noalign{\smallskip}\hline\noalign{\smallskip}
            $C_{frm}$    & .152          & .253          & .143          & .286      & .136      & .126      & .214      & .758      & .400      & .372 \\ 
            $C_{ask}$    & .286          & .237          & .359          & .244      & .356      & .107      & .310      & .582      & {\bf.579} & .280 \\
            $D_{tox}$    & .343          & .373          & .449          & .335      & .443      & .149      & .389      & .628      & .539 & \bf{.806} \\
            \noalign{\smallskip}\hline
        \end{tabular*}
\end{table}

Looking at Table~\ref{tab:base}, the upper group of rows under T1 represents the results for Experiment I. We posed in Hypothesis \textbf{1} that samples are underpowered regarding their representation of the language variation between platforms, both for bullying and normal language-use. The data analysis in Section~\ref{subs:sum} showed minimal overlap between domains in vocabulary and notable variances in numerous aspects of the available corpora. Consequently, we raised doubts regarding the ability of models trained on these individual corpora to generalize to other corpora (i.e., domains).

Firstly, we consider how well our \texttt{baseline} performed on the \emph{in-domain test sets}. For half of the corpora, it performs best overall on these specific sets (i.e., the test set portion of the data the model was trained on). More importantly, this entails that for four of the other sets, models trained on other corpora perform equal or better. Particularly the effectiveness of $D_{ask}$ was in some cases surprising; the YouTube corpus by \cite{dadvar2014experts} ($D_{ytb}$), for example, contains much longer instances (see Table~\ref{tab:dat}). 

It must be noted though, that the baseline was selected from work on the Ask.fm corpus \cite{vanhee2018}. This data is also one of the more diverse datasets (and largest) with exclusively short messages; therefore, one could assume a model trained on this data would work well on both longer and shorter instances. It is however also likely that particularly this baseline (binary word features) trained on this data therefore enforces the importance of more shallow features. This we will be further explored in Experiments II and III. 

\begin{figure*}
    \centering
    \begin{minipage}{0.74\textwidth}
        \begin{tikzpicture}
\pgfmathsetlengthmacro\MajorTickLength {
  \pgfkeysvalueof{/pgfplots/major tick length} * 0.4
}
\begin{axis}[
height=4cm,
width=12cm,
tick align=outside,
tick pos=left,
x grid style={white!69.01960784313725!black},
xmin=-0.5, xmax=19.5,
xtick style={color=black},
xtick={0,1,2,3,4,5,6,7,8,9,10,11,12,13,14,15,16,17,18,19},
xticklabel style = {rotate=45.0, xshift=-0.05cm, yshift=0.15cm},
xticklabels={b*tch,f*ck,f*gg*t,stfu,sl*t,g*y,c*nt,f*g,wh*r*,*sshole,n*gg*,hate,tw*t,d*ck,*ss,loser,ugly,go away,dumb*ss,f*ck*ng},
y grid style={white!69.01960784313725!black},
ymin=0, ymax=1.0,
ytick style={color=black},
major tick length=\MajorTickLength,
every tick label/.append style={font=\scriptsize}
]
\draw[fill=white!60.0!black,draw opacity=0] (axis cs:-0.25,0) rectangle (axis cs:0.25,0.486159664958781);
\draw[fill=white!60.0!black,draw opacity=0] (axis cs:0.75,0) rectangle (axis cs:1.25,0.397253179557668);
\draw[fill=white!60.0!black,draw opacity=0] (axis cs:1.75,0) rectangle (axis cs:2.25,0.391205443140379);
\draw[fill=white!60.0!black,draw opacity=0] (axis cs:2.75,0) rectangle (axis cs:3.25,0.384622402673902);
\draw[fill=white!60.0!black,draw opacity=0] (axis cs:3.75,0) rectangle (axis cs:4.25,0.376668541047829);
\draw[fill=white!60.0!black,draw opacity=0] (axis cs:4.75,0) rectangle (axis cs:5.25,0.361748956873954);
\draw[fill=white!60.0!black,draw opacity=0] (axis cs:5.75,0) rectangle (axis cs:6.25,0.355548188760811);
\draw[fill=white!60.0!black,draw opacity=0] (axis cs:6.75,0) rectangle (axis cs:7.25,0.348264270625347);
\draw[fill=white!60.0!black,draw opacity=0] (axis cs:7.75,0) rectangle (axis cs:8.25,0.342418077994587);
\draw[fill=white!60.0!black,draw opacity=0] (axis cs:8.75,0) rectangle (axis cs:9.25,0.321320944748114);
\draw[fill=white!60.0!black,draw opacity=0] (axis cs:9.75,0) rectangle (axis cs:10.25,0.319181918889777);
\draw[fill=white!60.0!black,draw opacity=0] (axis cs:10.75,0) rectangle (axis cs:11.25,0.28227852361097);
\draw[fill=white!60.0!black,draw opacity=0] (axis cs:11.75,0) rectangle (axis cs:12.25,0.272251151873976);
\draw[fill=white!60.0!black,draw opacity=0] (axis cs:12.75,0) rectangle (axis cs:13.25,0.269157546152352);
\draw[fill=white!60.0!black,draw opacity=0] (axis cs:13.75,0) rectangle (axis cs:14.25,0.268011651014183);
\draw[fill=white!60.0!black,draw opacity=0] (axis cs:14.75,0) rectangle (axis cs:15.25,0.248514820985584);
\draw[fill=white!60.0!black,draw opacity=0] (axis cs:15.75,0) rectangle (axis cs:16.25,0.245139144071666);
\draw[fill=white!60.0!black,draw opacity=0] (axis cs:16.75,0) rectangle (axis cs:17.25,0.24213476984973);
\draw[fill=white!60.0!black,draw opacity=0] (axis cs:17.75,0) rectangle (axis cs:18.25,0.241434672697737);
\draw[fill=white!60.0!black,draw opacity=0] (axis cs:18.75,0) rectangle (axis cs:19.25,0.240633424099152);
\path [draw=black, semithick]
(axis cs:0,0.0560842165069208)
--(axis cs:0,0.916235113410641);

\path [draw=black, semithick]
(axis cs:1,0.0955653650260613)
--(axis cs:1,0.698940994089275);

\path [draw=black, semithick]
(axis cs:2,0.0220664643172784)
--(axis cs:2,0.760344421963479);

\path [draw=black, semithick]
(axis cs:3,0.000380341105266302)
--(axis cs:3,0.768864464242538);

\path [draw=black, semithick]
(axis cs:4,-0.0116548788890098)
--(axis cs:4,0.764991960984667);

\path [draw=black, semithick]
(axis cs:5,-0.0927967613667052)
--(axis cs:5,0.816294675114612);

\path [draw=black, semithick]
(axis cs:6,-0.0854321211301985)
--(axis cs:6,0.796528498651821);

\path [draw=black, semithick]
(axis cs:7,-0.0956444615248115)
--(axis cs:7,0.792173002775506);

\path [draw=black, semithick]
(axis cs:8,-0.0979864792153251)
--(axis cs:8,0.782822635204499);

\path [draw=black, semithick]
(axis cs:9,-0.0120080354445238)
--(axis cs:9,0.654649924940752);

\path [draw=black, semithick]
(axis cs:10,0.0176483956400034)
--(axis cs:10,0.620715442139551);

\path [draw=black, semithick]
(axis cs:11,-0.0181840890003955)
--(axis cs:11,0.582741136222336);

\path [draw=black, semithick]
(axis cs:12,-0.072987181681835)
--(axis cs:12,0.617489485429786);

\path [draw=black, semithick]
(axis cs:13,-0.120224980010959)
--(axis cs:13,0.658540072315663);

\path [draw=black, semithick]
(axis cs:14,-0.111389098729857)
--(axis cs:14,0.647412400758223);

\path [draw=black, semithick]
(axis cs:15,-0.0415683394533368)
--(axis cs:15,0.538597981424504);

\path [draw=black, semithick]
(axis cs:16,-0.0021349911412078)
--(axis cs:16,0.49241327928454);

\path [draw=black, semithick]
(axis cs:17,-0.0318218411882086)
--(axis cs:17,0.516091380887669);

\path [draw=black, semithick]
(axis cs:18,-0.0250316041955573)
--(axis cs:18,0.507900949591032);

\path [draw=black, semithick]
(axis cs:19,-0.0437376242413434)
--(axis cs:19,0.525004472439648);

\draw[] (axis cs:0.1,0.586159664958781) -- (axis cs:0.1,0.586159664958781);
\node at (axis cs:0.1,0.586159664958781)[
  scale=0.5,
  anchor= west,
  text=black,
  rotate=0.0
]{1141};
\draw[] (axis cs:1.1,0.497253179557668) -- (axis cs:1.1,0.497253179557668);
\node at (axis cs:1.1,0.497253179557668)[
  scale=0.5,
  anchor= west,
  text=black,
  rotate=0.0
]{2685};
\draw[] (axis cs:2.1,0.491205443140379) -- (axis cs:2.1,0.491205443140379);
\node at (axis cs:2.1,0.491205443140379)[
  scale=0.5,
  anchor= west,
  text=black,
  rotate=0.0
]{11};
\draw[] (axis cs:3.1,0.484622402673902) -- (axis cs:3.1,0.484622402673902);
\node at (axis cs:3.1,0.484622402673902)[
  scale=0.5,
  anchor= west,
  text=black,
  rotate=0.0
]{247};
\draw[] (axis cs:4.1,0.476668541047829) -- (axis cs:4.1,0.476668541047829);
\node at (axis cs:4.1,0.476668541047829)[
  scale=0.5,
  anchor= west,
  text=black,
  rotate=0.0
]{119};
\draw[] (axis cs:5.1,0.461748956873954) -- (axis cs:5.1,0.461748956873954);
\node at (axis cs:5.1,0.461748956873954)[
  scale=0.5,
  anchor= west,
  text=black,
  rotate=0.0
]{11};
\draw[] (axis cs:6.1,0.455548188760811) -- (axis cs:6.1,0.455548188760811);
\node at (axis cs:6.1,0.455548188760811)[
  scale=0.5,
  anchor= west,
  text=black,
  rotate=0.0
]{130};
\draw[] (axis cs:7.1,0.448264270625347) -- (axis cs:7.1,0.448264270625347);
\node at (axis cs:7.1,0.448264270625347)[
  scale=0.5,
  anchor= west,
  text=black,
  rotate=0.0
]{409};
\draw[] (axis cs:8.1,0.442418077994587) -- (axis cs:8.1,0.442418077994587);
\node at (axis cs:8.1,0.442418077994587)[
  scale=0.5,
  anchor= west,
  text=black,
  rotate=0.0
]{279};
\draw[] (axis cs:9.1,0.421320944748114) -- (axis cs:9.1,0.421320944748114);
\node at (axis cs:9.1,0.421320944748114)[
  scale=0.5,
  anchor= west,
  text=black,
  rotate=0.0
]{202};
\draw[] (axis cs:10.1,0.419181918889777) -- (axis cs:10.1,0.419181918889777);
\node at (axis cs:10.1,0.419181918889777)[
  scale=0.5,
  anchor= west,
  text=black,
  rotate=0.0
]{621};
\draw[] (axis cs:11.1,0.38227852361097) -- (axis cs:11.1,0.38227852361097);
\node at (axis cs:11.1,0.38227852361097)[
  scale=0.5,
  anchor= west,
  text=black,
  rotate=0.0
]{1082};
\draw[] (axis cs:12.1,0.372251151873976) -- (axis cs:12.1,0.372251151873976);
\node at (axis cs:12.1,0.372251151873976)[
  scale=0.5,
  anchor= west,
  text=black,
  rotate=0.0
]{65};
\draw[] (axis cs:13.1,0.369157546152352) -- (axis cs:13.1,0.369157546152352);
\node at (axis cs:13.1,0.369157546152352)[
  scale=0.5,
  anchor= west,
  text=black,
  rotate=0.0
]{182};
\draw[] (axis cs:14.1,0.368011651014183) -- (axis cs:14.1,0.368011651014183);
\node at (axis cs:14.1,0.368011651014183)[
  scale=0.5,
  anchor= west,
  text=black,
  rotate=0.0
]{125};
\draw[] (axis cs:15.1,0.348514820985584) -- (axis cs:15.1,0.348514820985584);
\node at (axis cs:15.1,0.348514820985584)[
  scale=0.5,
  anchor= west,
  text=black,
  rotate=0.0
]{499};
\draw[] (axis cs:16.1,0.345139144071666) -- (axis cs:16.1,0.345139144071666);
\node at (axis cs:16.1,0.345139144071666)[
  scale=0.5,
  anchor= west,
  text=black,
  rotate=0.0
]{564};
\draw[] (axis cs:17.1,0.34213476984973) -- (axis cs:17.1,0.34213476984973);
\node at (axis cs:17.1,0.34213476984973)[
  scale=0.5,
  anchor= west,
  text=black,
  rotate=0.0
]{426};
\draw[] (axis cs:18.1,0.341434672697737) -- (axis cs:18.1,0.341434672697737);
\node at (axis cs:18.1,0.341434672697737)[
  scale=0.5,
  anchor= west,
  text=black,
  rotate=0.0
]{82};
\draw[] (axis cs:19.1,0.340633424099152) -- (axis cs:19.1,0.340633424099152);
\node at (axis cs:19.1,0.340633424099152)[
  scale=0.5,
  anchor= west,
  text=black,
  rotate=0.0
]{197};
\end{axis}

\end{tikzpicture}
    \end{minipage} \hfill
    \begin{minipage}{0.23\textwidth}
        \begin{tikzpicture}
\pgfmathsetlengthmacro\MajorTickLength {
  \pgfkeysvalueof{/pgfplots/major tick length} * 0.4
}
\begin{axis}[
width=4.3cm,
height=4cm,
log basis y={10},
tick align=outside,
tick pos=left,
x grid style={white!69.01960784313725!black},
xmin=-0.5, xmax=5.5,
xtick style={color=black},
xtick={0,1,2,3,4,5},
xlabel={\strut},
xticklabel style = {rotate=0.0},
xticklabels={1,2,3,4,5,6},
y grid style={white!69.01960784313725!black},
ymin=198.272206672272, ymax=2383.78342548652,
ytick style={color=black},
major tick length=\MajorTickLength,
every tick label/.append style={font=\scriptsize}
]
\draw[fill=white!60.0!black,draw opacity=0] (axis cs:-0.25,0) rectangle (axis cs:0.25,2129);

\draw[fill=white!60.0!black,draw opacity=0] (axis cs:0.75,0) rectangle (axis cs:1.25,1296);
\draw[fill=white!60.0!black,draw opacity=0] (axis cs:1.75,0) rectangle (axis cs:2.25,864);
\draw[fill=white!60.0!black,draw opacity=0] (axis cs:2.75,0) rectangle (axis cs:3.25,235);
\draw[fill=white!60.0!black,draw opacity=0] (axis cs:3.75,0) rectangle (axis cs:4.25,222);
\draw[fill=white!60.0!black,draw opacity=0] (axis cs:4.75,0) rectangle (axis cs:5.25,254);
\draw[] (axis cs:0,2129) -- (axis cs:0,2129);
\node at (axis cs:0,2129)[
  scale=0.5,
  anchor=south west,
  text=black,
  rotate=0.0
]{2129 (25.5\%)};
\draw[] (axis cs:1,1296) -- (axis cs:1,1296);
\node at (axis cs:1,1296)[
  scale=0.5,
  anchor=south west,
  text=black,
  rotate=0.0
]{1296 (15.5\%)};
\draw[] (axis cs:2,864) -- (axis cs:2,864);
\node at (axis cs:2,864)[
  scale=0.5,
  anchor=south west,
  text=black,
  rotate=0.0
]{864 (10.3\%)};
\draw[] (axis cs:3,235) -- (axis cs:3,235);
\node at (axis cs:3,235)[
  scale=0.5,
  anchor=south west,
  text=black,
  rotate=0.0
]{235};
\draw[] (axis cs:4,222) -- (axis cs:4,222);
\node at (axis cs:4,222)[
  scale=0.5,
  anchor=south west,
  text=black,
  rotate=0.0
]{222};
\draw[] (axis cs:5,254) -- (axis cs:5,254);
\node at (axis cs:5,254)[
  scale=0.5,
  anchor=south west,
  text=black,
  rotate=0.0
]{254};
\end{axis}

\end{tikzpicture}
    \end{minipage}
    \caption{Left: Top 20 \emph{test set} words with the highest average coefficient values across all classifiers (minus the model trained on $D_{tox}$). Error bars represent standard deviation. Each coefficient value is only counted once per test set. The frequency of the words is listed in the annotation. Right: Test set occurrence frequencies (and percentages) of the top 5,000 highest absolute feature coefficient values.}
    \label{fig:bars}
\end{figure*}
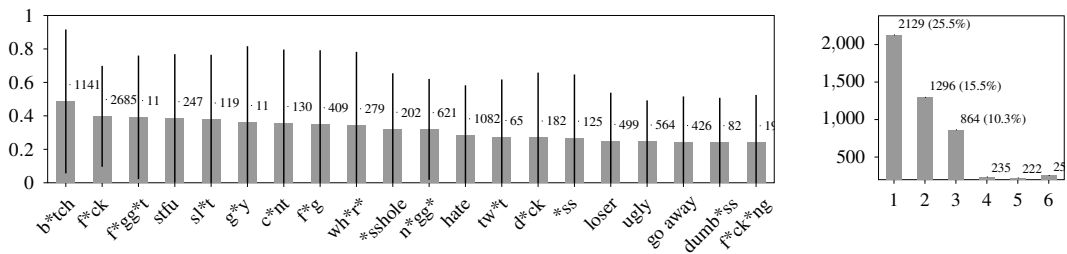

For Experiment I, however, our goal was to assess the out-of-domain performance of these classifiers, not to maximize performance. For this, we turn to the \emph{Avg} column in Table~\ref{tab:base}. Between the top portion of the Table, the $D_{ask}$ model performs best across all domains (achieving highest on three, as mentioned above). The second-best model is trained on the Formspring data from \cite{Reynolds2011} ($D_{frm}$), akin to Ask.fm as a domain (question-answer style, option to post anonymously). It can be observed that almost all models perform worst on the `bullying traces' Twitter corpus by \cite{Xu2012}, which was collected using queries. This result is relatively unsurprising, given the small vocabulary overlaps with its test set shown in Figure~\ref{fig:jac}. We also confirm in line with \cite{Reynolds2011} that the \textsc{caw} data from \cite{Bayzick2011} is unfit as a bullying corpus; achieving significant positive $F_1$-scores with a baseline, generalizing poorly and proving difficult as a test set.

Additionally, we observe that even the best performing models yield between $.1$ and $.2$ lower $F_1$ scores on other domains, or a $15-30\%$ drop from the original score. To explain this, we look at how well important features generalize across test sets. As our \texttt{baseline} is a Linear SVM, we can directly extract all grams with positive coefficients (i.e., related to bullying). Figure~\ref{fig:bars} (right) shows the frequency of the top 5,000 features with the highest coefficient values. These can be observed to follow a Zipfian-like distribution, where the important features most frequently occur in one test set ($25.5\%$) only, which quickly drops off with increasing frequency. Conversely, this implies that over $75\%$ of the top 5,000 features seen during training do not occur in any test instance, and only $3\%$ generalize across all sets. This coverage decreases to roughly $60\%$ and $4\%$ respectively for the top 10,000, providing further evidence of the strong variation in predominantly bullying-specific language-use.

Figure~\ref{fig:bars} (left) also indicates that the coefficient values are highly unstable across test sets, with most having roughly a $0.4$ standard deviation. Note that these coefficient values can also flip to negative for particular sets, so for some of the features, the range goes from associated with the other class to highly associated with bullying. Given the results of Table~\ref{tab:base} and Figure~\ref{fig:bars}, we can conclude that our \texttt{baseline} model shows not to generalize out-of-domain. Given the quantitative and qualitative results reported on in this Experiment, this particular setting partly supports Hypothesis \textbf{1}.

\subsection{Experiment II}

\begin{table} \small
    \caption{Examples of uni-gram weights according to the \texttt{baseline} SVM trained $D_{all}$, tested on $D_{twB}$ and $D_{ask}$. Words in red are associated with bullying, words in green with neutral content. The color intensity is derived from the strength of the SVM coefficients per feature (most are near zero). Black boxes indicate OOV words. Labels are divided between the gold standard ($y$) and predicted ($\hat{y}$) labels, \faThumbsODown{} for bullying content, \faThumbsOUp{} for neutral. }
    \label{tab:qual}
    \begin{tabular*}{\textwidth}{@{\extracolsep{\fill}} llp{5cm}p{5cm}}
        \hline\noalign{\smallskip}
        $y$             & $\hat{y}$         &  $D_{twB}$     &  $D_{ask}$  \\
        \noalign{\smallskip}\hline\noalign{\smallskip}
        \faThumbsOUp    & \faThumbsODown    & \colorbox{pc!1!}{about} \colorbox{pc!3!}{to} \colorbox{nc!42!}{leave} \colorbox{nc!4!}{this} \colorbox{pc!7!}{school} \colorbox{pc!1!}{library} \colorbox{pc!1!}{and} \colorbox{pc!3!}{take} \colorbox{nc!3!}{my} \colorbox{nc!62!}{*ss} \framebox{homeeee} & bigerrr ? \colorbox{pc!2!}{how} \colorbox{nc!3!}{much} ? \colorbox{pc!3!}{its} \colorbox{nc!4!}{gon} \colorbox{pc!4!}{na} \colorbox{nc!15!}{touch} \colorbox{pc!2!}{the} \colorbox{pc!5!}{sky} \colorbox{nc!0!}{?} a \colorbox{pc!5!}{wonder} \colorbox{nc!84!}{d*ck} \colorbox{nc!0!}{?}  \\
        \faThumbsODown   & \faThumbsODown   & \colorbox{nc!19!}{you} \colorbox{nc!63!}{p*ss} \colorbox{nc!8!}{me} \colorbox{nc!28!}{off} \colorbox{nc!1!}{so} \colorbox{nc!3!}{much} . & \colorbox{nc!0!}{r} \colorbox{nc!0!}{u} \colorbox{nc!0!}{a} \colorbox{nc!62!}{r*t*rd} \colorbox{nc!3!}{liam} \colorbox{nc!20!}{mate} \colorbox{nc!94!}{f*ck} \colorbox{nc!28!}{off}  \\
        \faThumbsODown   & \faThumbsOUp     & \framebox{@username} \colorbox{nc!0!}{i} \colorbox{nc!6!}{will} \colorbox{nc!11!}{skull} \colorbox{nc!2!}{drag} \colorbox{nc!19!}{you} \colorbox{pc!2!}{across} \colorbox{nc!1!}{campus} . &  \colorbox{nc!1!}{h*} \colorbox{nc!1!}{of} \colorbox{nc!1!}{me} \colorbox{nc!3!}{xoxoxoxoxoxoox} \\
        \noalign{\smallskip}\hline
    \end{tabular*}
\end{table}

The results for this experiment can be predominantly found in Table~\ref{tab:base} (middle and lower parts, and T2 in particular), and partly in Table~\ref{tab:archs} (word2vec, DistilBERT). In this experiment, we seek to further test Hypothesis \textbf{1} by employing three methods: merging all cyberbullying data to increase \emph{volume and variety}, aggregating on context level for a \emph{context change}, and \emph{improving representations} through pre-trained word embedding features. These are all reasonably straightforward methods that can be employed in an attempt to mitigate data scarcity.

\paragraph{Volume and Variety} The results for this part are listed under $D_{all}$ in Table~\ref{tab:base}. For all of the following experiments, we now focus on the full results table (including that of Experiment I) and see which individual classifiers generalize best across all test sets (highlighted in gray). The \emph{Avg} column shows that our `big' model trained on all available corpora\footnote{This average excludes toxicity data from $D_{tox}$, which we found when added to substantially decrease performance on all domains, except for $D_{twX}$ and $C_{frm}$. Note that it also includes scopes from the \emph{context change} experiment.} achieves second-best performance on half of the test sets and best on the other half. More importantly, it has the highest average out-of-domain performance, without competition on any test set. These observations imply that for the \texttt{baseline} setting, an ensemble model of different smaller classifiers should not be preferred over the big model. Consequently, it can be concluded that collecting more data does seem to aid the task as a whole.

However, a qualitative analysis of the predictions made by this model clearly shows lingering limitations (see Table~\ref{tab:qual}). These three randomly-picked examples give a clear indication of the focus on blatant profanity (such as \texttt{d*ck}, \texttt{p*ss}, and \texttt{f*ck}). Especially combinations of words that in isolation might be associated with bullying content (\texttt{leave}, \texttt{touch}) tend to confuse the model. It also fails to capture more subtle threats (\texttt{skull drag}) and infrequent variations (\texttt{h*}). Both of these structural mistakes could be mitigated by providing more context that potentially includes either more toxicity or more examples of neutral content to decrease the impact of single curse words---hence, the next experiment.

\paragraph{Context Change} As for access to context scopes, we are restricted to the Ask.fm and Formspring data ($C_{frm}$ and $C_{ask}$ in Table~\ref{tab:base}). Nevertheless, in both cases, we see a noticeable increase for in-domain performance: a positive $F_1$ score of .579 for context scope versus .561 on Ask.fm, and .758 versus .454 on Formspring respectively. This increase implies that considering message-level detection for both individual sets should be preferred. On the other hand, however, these longer contexts do perform worse on out-of-domain sets.

On manual inspection of the feature differences between the other sets, $D_{ask}$ and $D_{frm}$ individually, and $C_{frm}$ and $C_{ask}$, the scope shift clearly shows in their importances. From a sample of 500 top features occurring in the test set,  63\% are profane words. For the models trained on Ask.fm and Formspring this is an average of 42\%, and models trained on both context scopes, it is significantly reduced to 11\%. Many important bi-gram features include \texttt{you}, topics such as \texttt{dating}, \texttt{boys}, \texttt{girls}, and \texttt{girlfriend} occur, yet also positive words such as (\texttt{are}) \texttt{beautiful}---the latter of which could indicate messages from friends (defenders). This change is to an extent expected as by changing the scope, the task shifts to classifying profiles that are bullied, thus showing more diverse bullying characteristics.

These results provide evidence for extending classification to contexts to be a worthwhile platform-specific setting to pursue. However, we can conversely draw the same conclusions as Experiment I; that including direct context does not overcome the tasks general domain limitations, therefore further supporting Hypothesis \textbf{1}. A plausible solution to this could be improving upon the BoW features by relying on more general representations of language, as found in word embeddings.

\paragraph{Improving Representations} 

\begin{table} \small
        \caption{Overview of different feature representations (Repr) for Experiment I and II. The `+' parts show performance for preprocessing: removing all special characters (clean), and more sophisticated handling of social media tags and emojis (preproc). Their in-domain positive class $F_1$ scores for Experiment I (T1) and II (T2), and the out-of-domain average (Avg) for $D_{all}$. Baseline scores are from Table~\ref{tab:base}.}\label{tab:archs}
        \begin{tabular*}{\textwidth}{@{\extracolsep{\fill}} lrrrrrrrrrr}
            \hline\noalign{\smallskip}
            Repr          &                             \multicolumn{6}{c}{T1}                             & Avg        & \multicolumn{2}{c}{T2} & T3 \\
                          \cmidrule{2-7} \cmidrule{8-8} \cmidrule{9-10} \cmidrule{11-11}
                          & $D_{twB}$ & $D_{frm}$    & $D_{msp}$   & $D_{ytb}$ & $D_{ask}$  & $D_{twX}$    &            & $C_{frm}$ & $C_{ask}$ & $D_{tox}$  \\ 
            \noalign{\smallskip}\hline\noalign{\smallskip}
            {\tt baseline}  & \bf{.417}    & .454        & \bf{.941}  & .365      & .561      & .508       & .557      & .758      & .579      & .806 \\
            + clean         & .408         & \bf{.477}   & .927       & .354      & \bf{.562} & \bf{.517}  & .561      & \bf{.764} & .592      & .\bf{807} \\
            + preproc       & .345         & .426        & .929       & \bf{.377} & .506      & .293       & .512      & .600      & .582      & .734 \\
            \noalign{\smallskip}\hline\noalign{\smallskip}
            NBSVM           & .364         & .462        & .929       & .231      & .508      & .469       & .542      & .635      & .592      & .779\\
            + clean         & .410         & .456        & .940       & .211      & .541      & .467       & .563      & .641      & .596      & .747 \\
            + preproc       & .318         & .466        & .907       & .320      & .480      & .305       & \bf{.566} & .532      & .597 & .756 \\
            \noalign{\smallskip}\hline\noalign{\smallskip}
            word2vec        & .368         & .394        & .860       & .338      & .304      & .323       & .366      & .698      & .572      & .634 \\
            DistilBERT      & .377         & .336        & .697       & .296      & .369      & .435       & .402      & .598      & \bf{.629}      & .642 \\
            \noalign{\smallskip}\hline
        \end{tabular*}
\end{table}

The aim for this experiment was to find (out-of-the-box) representations that would improve upon the simple BoW features used in our \texttt{baseline} model (i.e., achieving good in-domain performance as well as out-of-domain generalization). Table~\ref{tab:archs} lists both of our considered baselines, tested under different preprocessing methods. These are then compared against the two different embedding representations. 

For preprocessing, several levels were used: the default for all models being 1) lowercasing only, then either 2) removal of special characters, or 3) lemmatization and more appropriate handling of special characters (e.g., splitting \texttt{\#word} to prepend a hashtag token) were added. The corresponding results in Table~\ref{tab:archs} do not reveal an unequivocal preprocessing method for either the BoW \texttt{baseline} or NBSVM. While the latter achieves highest out-of-domain generalization with thorough preprocessing (`+preproc', .566 positive $F_1$), the \texttt{baseline} model achieves best in-domain performance on five out of nine corpora, and an on-par out-of-domain average ($.566$ versus $.561$) with simple cleaning (`+clean').

According to our criterion proposed in Section~\ref{subs:crit}, the method with good in- and out-of-domain should be preferred. The current consideration of preprocessing methods illustrates how this stricter evaluation criterion used in this experiment potentially yields different overall results in contrast to evaluating in-domain only, or focusing on single corpora. Conversely, we opted for simple cleaning throughout the rest of our experiment (as mentioned in Section~\ref{subs:proc}), given its consistent performance for both baselines. 

The embeddings do not seem to provide representations that yield and overall improvement for the classification performance of our Logistic Regression model. Surprisingly, however, DistilBERT does yield significant gains over our baseline for the conversation-level corpus of Ask.fm ($.629$ positive $F_1$ over $.579$). This might imply that such representations would work well on more (balanced) data, although fine-tuning would be a requirement for drawing strong conclusions. Moreover, given that we restricted our embeddings to averaged representations on document-level for word2vec, and the sentence representation token for BERT, other settings remain unexplored; however, are not in scope of the current work. Therefore, we can conclude that no other alternative (out-of-the-box) baselines seem to clearly outperform our BoW baseline. We previously eluded to its effectiveness in previous work, and argued this being a result of capturing blatant profanity. We will further test this in the next experiment.

\subsection{Experiment III}

Here, we investigate Hypothesis \textbf{2}: the notion that positive instances across all cyberbullying corpora are biased, and only reflect a limited dimension of bullying. We have already found strong evidence for this in the previous Experiments I and II, Figure~\ref{fig:bars}, Table~\ref{tab:qual}, and manual analyses of top features all indicated toxicity to be consistent top-ranking features. To add more empirical evidence to this, we trained models on toxicity, or cyber aggression, and tested them on bullying data (and vice-versa)---providing results on the overlap between the tasks. The results for this experiment can be found in the lower end of Table~\ref{tab:base}, under $D_{tox}$ and T3.

It can be noted that there is a substantial gap in performance between the cyberbullying classifiers (using $D_{all}$ as reference) performance on the $D_{tox}$ test set and that of the toxicity model (positive $F_1$ score of $.587$ and $.806$ respectively). More strikingly, however, the other way around, toxicity classifiers perform second-best on the out-of-domain averages (\emph{Avg} in Table~\ref{tab:base}). In the context scopes ($C_{frm}$ and $C_{ask}$) it is notably close, and for other sets relatively close, to the in-domain performance.

 Cyberbullying detection should include detection of toxic content, yet also perform on more complex social phenomena, likely not found in the Wikipedia comments of the toxicity corpus. It is therefore particularly surprising that it achieves higher out-of-domain performance on cyberbullying classification than all individual models using BoW features to capture bullying content. Only when all corpora are combined, the $D_{all}$ classifier performs better than the toxicity model. This observation combined with previous results provides significant evidence that a large part of the available cyberbullying content is not complex, and current models to only generalize to a limited extent using predominately simple aggressive features, supporting Hypothesis \textbf{3}.

\subsection{Experiment IV}

\begin{table} \small
        \caption{Overview of different architectures (Arch) their in-domain positive class $F_1$ scores for Experiment I (T1) and II (T2), the out-of-domain average for $D_{all}$ ($all$), and $D_{tox}$ ($tox$). Baseline model (and scores) is that of Table~\ref{tab:base}. Reproduction results of \cite{agrawal2018deep} are denoted by *, their oversampling method by +. Our tuned model versions have no annotation, character level models are denoted by $\star$.}
        \label{tab:neural}
        \begin{tabular*}{\textwidth}{@{\extracolsep{\fill}} lrrrrrrrrrr}
            \hline\noalign{\smallskip}
            Arch          &                             \multicolumn{6}{c}{T1}                         & Avg  &   \multicolumn{2}{c}{T2} & T3 \\
                          \cmidrule{2-7} \cmidrule{8-8} \cmidrule{9-10} \cmidrule{11-11}
                          & $D_{twB}$ & $D_{frm}$    & $D_{msp}$   & $D_{ytb}$ & $D_{ask}$  & $D_{twX}$  & $all$ \hspace{0.1cm} $tox$ & $C_{frm}$ & $C_{ask}$ & $D_{tox}$  \\ 
            \noalign{\smallskip}\hline\noalign{\smallskip}
            {\tt baseline}  & .417         & .454        & .941       & .365      & \bf{.561} & \bf{.508}  & \bf{.557} \hspace{0.1cm} \bf{.389}   & \bf{.758} & .579      & \bf{.806} \\
            NBSVM           & .383         & \bf{.486}   & .925       & \bf{.387} & .476      & .396       & .551 \hspace{0.1cm} .385             & .703      & .604      & .797 \\
            \noalign{\smallskip}\hline\noalign{\smallskip}
            BiLSTM*         & .171         & .363        & .938       & .152      & .504      & .400       & .440 \hspace{0.1cm} .349             & .609      & .507      & .762 \\
            BiLSTM+         & .188         & .396        & \bf{.951}  & .160      & .438      & .341       & .417 \hspace{0.1cm} .337             & .541      & .505      & .737 \\
            BiLSTM          & .182         & .341        & .905       & .148      & .463      & .246       & .479 \hspace{0.1cm} .356             & .608      & .522      & .774 \\
            \noalign{\smallskip}\hline\noalign{\smallskip}
            CNN*            & \bf{.500}    & .276        & .790       & .133      & .462      & .438       & .364 \hspace{0.1cm} .350             & .000      & .306      & .753 \\
            CNN             & .444         & .416        & .816       & .000      & .498      & .438       & .464 \hspace{0.1cm} .342             & .000      & .610      & .754 \\ 
            CNN$\star$      & .444         & .419        & .816       & .000      & .499      & .375       & .460 \hspace{0.1cm} .362             & .000      & \bf{.647} & .774 \\ 
            C-LSTM*         & .000         & .421        & .875       & .095      & .000      & .000       & .449 \hspace{0.1cm} .329             & .094      & .425      & .757 \\
            C-LSTM          & .000         & .019        & .829       & .000      & .066      & .000       & .463 \hspace{0.1cm} .355             & .095      & .518      & .761 \\
            C-LSTM$\star$   & .000         & .057        & .853       & .075      & .008      & .000       & .278 \hspace{0.1cm} .358             & .296      & .506      & .756 \\
            \noalign{\smallskip}\hline
        \end{tabular*}
\end{table}

So far, we have attempted to improve a straight-forward baseline that was trained on binary features with several different approaches. While changes in data (representations) seem to have a noticeable effect on performance (increasing the amount of messages per instance, merging all corpora), none of the experiments with different feature representations have had an impact. With the current experiment, we had hoped to leverage earlier state-of-the-art architectures by reproducing their methodology and subjecting  our evaluation framework.

As can be inferred from Table~\ref{tab:neural}, our baselines outperform these neural techniques on almost all in-domain tests, as well as the out-of-domain averages. Having strictly upheld the experimental set-up from \cite{agrawal2018deep} and as close as possible that of \cite{rosa2018deeper}, we can conclude that---under stricter evaluation---there is sufficient evidence that these models do not provide state-of-the-art results on the task of cyberbullying.\footnote{Upon acquiring the results of the replication of \cite{agrawal2018deep} (in particular failing to replicate the effect of the paper's oversampling) we investigated the provided code and notebooks. It is our understanding that oversampling before splitting the dataset into training and test sets causes the increase in performance; we measured overlap of positive instances in these splits and found no unique test instances. Furthermore, after re-running the experiments directly from the notebooks with the oversampling conducted post-split, the effect was significantly decreased (similar to our results in Table~\ref{tab:neural}). The authors were contacted with our observations in March 2019. They unfortunately have not yet confirmed our results. Their repository remains unchanged as of October 2019. Our analyses can be found here:   \url{https://github.com/cmry/amica/tree/master/reproduction}.} Tuning these networks (at least in our set-up) does not seem to improve performance, rather decrease it. This indicates that the validation set on which early stopping is conducted is often not representative to the test set. Parameter tuning on this set is consequently sensitive to overfitting; an arguably unsurprising result given the size of the corpora.

Some further noteworthy observations can be made related to the performance of the CNN architecture, achieving quite significant leaps on word level (for $D_{twB}$) and character level (for $C_{ask}$). Particularly the conversation scopes ($C$, with a comparitatively balanced class distribution) see much more competitive perfomance compared to the baselines. The same effect can be observed when more data is available; both averages test scores for $D_{all}$ and $D_{tox}$ are comparable to the baseline across almost all architectures. Additionally, the $D_{tox}$ scores indicate that all architectures show about the same overlap on toxicity detection, although interestingly, less so for the neural models than for the baselines. 

It can therefore be concluded that the current neural architectures do not provide a solution to the limitations of the task, rather, suffering more in performance. Our experiments do, however, once more illustrate that the proposed techniques of improving the representations of the corpora (by providing more data through merging all sources, and balancing by classifying batches of multiple messages, or conversations) allow the neural models to approach the baseline ballpark. As our goal here was not to completely optimize these architectures, but replication, the proposed techniques still could provide more avenues for further research. Finally, given its robust performance, we will continue to use the baseline model for the next eperiment.

\subsection{Experiment V}

\begin{table}[t!] \small
    \caption{Positive class $F_1$ scores for Experiment IV on Dutch data. Models are fitted on the training proportion of the corpora row-wise and tested column-wise. The best overall test score is noted in bold. The scores of primary interest are highlighted in gray.} \label{tab:basenl}
    \begin{tabular*}{\textwidth}{@{\extracolsep{\fill}} lrrrrr}
        \hline\noalign{\smallskip}
        train               & \multicolumn{3}{c}{T1}                      & \multicolumn{2}{c}{T2}         \\
                            \cmidrule{2-4} \cmidrule{5-6}
                            & $D_{ask\_nl}$ & $D_{sim\_nl}$ & $D_{don\_nl}$   &  $D_{askC\_nl}$    & $D_{simC\_nl}$ \\ 
        \noalign{\smallskip}\hline\noalign{\smallskip}
        $D_{ask\_nl}$       & .598          & .516          & .495            & .264             & .533             \\
        $D_{sim\_nl}$       & .273          & \bf{.708}     & \gc{\bf.667}    & .501             & .800             \\
        \noalign{\smallskip}\hline\noalign{\smallskip}
        $D_{comb}$          & \gc{\bf.608}  & .681          & .516            & \gc{\bf.801}     & .808             \\
        \noalign{\smallskip}\hline\noalign{\smallskip}
        $D_{askC\_nl}$      & .165          & .361          & .182            & .505             & .750             \\
        $D_{simC\_nl}$      & .175          & .424          & .333            & .496             & .750             \\
        \noalign{\smallskip}\hline\noalign{\smallskip}
        $D_{all\_nl}$       & .577          & .677          & .516            & .379             & \bf{.821}        \\
        \noalign{\smallskip}\hline
    \end{tabular*}

\end{table}

Due to the nature of its experimental set-up (which generates balanced data with simple language-use, as shown in Table~\ref{tab:dat}), the crowdsourced data proves easy to classify. Therefore, we do not report out-of-domain averages, as this set would skew them too optimistically, and be uninformative. Regardless, we are primarily interested in performance when crowdsourced data is added, or used as a replacement for real data. In contrast to the other experiments, the focus will mostly be on the Ask.fm ($D_{ask\_nl}$) and donated ($D_{don\_nl}$) scores (see Table~\ref{tab:basenl}). The scores on the Dutch part of the Ask.fm corpus are quite similar to those on the English corpus ($.561$ vs $.598$ positive $F_1$ score), which is line with earlier results \cite{vanhee2018}. Moreover, particularly for the small amount of data, the crowdsourced corpus performs surprisingly well on $D_{ask\_nl}$ ($.516$), and significantly better on the donated test data ($.667$ on $D_{don\_nl}$).

In the settings that utilize context representations, training on conversation scopes initially does not seem to improve detection performance in any of the configurations (save for a  marginal gain on $D_{simC\_nl}$). However, it does simplify the task in a meaningful way at  test-time; whereas a slight gain is obtained for message-level $D_{ask\_nl}$ (from $.598$ $F_1$-score to $.608$), when merging both datasets a significant performance boost can be found when training on $D_{comb}$ and testing on $D_{askC\_nl}$ (from $.264$ and $.501$ to $.801$ on the combined). Hence, it can be concluded that enriching the existing training set with crowdsourced data yields promising improvements.

Based on these results, we confirm the Experiment II results hold for Dutch: more diverse, larger datasets, and increasing context sizes contributes to better performance on the task. Most importantly, there is enough evidence to support Hypothesis \textbf{3}: the data generated by the crowdsourcing experiment helps detection rates for our in-the-wild test set, and its combination with externally collected data increases performance with and without additional context.

\subsection{Suggestions for Future Work}

We hope our experiments have helped shed light, and will raise further attention regarding multiple issues with methodological rigor pertaining the task of cyberbullying detection. It is our understanding that the disproportionate amount of work on the (oversimplified) classification task, versus the lack of focus on constructing rich, representative corpora reflecting the actual dynamics of bullying, has made critical assessment of the advances in this task difficult. We would therefore want to particularly stress the importance of simple baselines and the out-of-domain tests that we included in the evaluation criterion for this research. They would provide a fairer comparison for proposed novel classifiers, and a more unified method of evaluation.

Furthermore, novel research would benefit from explicitly finding evidence to support its assumptions that classifiers labeled `cyberbullying detection' do more than one-shot, message-level toxicity detection. We would argue that the current framing of the majority of work on the task is still too limited to be considered theoretically-defined cyberbullying classification. In our research, we demonstrated several qualitative and quantitative methods that can facilitate such analyses. As popularity of the application of cyberbullying detection is increasing, this would avoid misrepresenting the conducted work, and that of possible in-the-wild applications in the future. 

While we demonstrated a method of collecting plausible cyberbullying with guaranteed consent, the more valuable sources of real-life bullying that allow for complex models of social interaction remain restricted. It is our expectation that future modeling will benefit from the construction of much larger (anonymized) corpora---as most fields dealing with language have, and we therefore hope to see future work heading this direction.

\section{Conclusion}

In this work, we identified several issues that affect the majority of the current research on cyberbullying detection. As it is difficult to collect accurate cyberbullying data in the wild, the field suffers from data scarcity. In an optimal scenario, rich representations capturing all required meta-data to model the complex social dynamics of what the literature defines as cyberbullying would likely prove fruitful. However, one can assume such access to remain restricted for the time being, and with current social media moving towards private communication, to not be generalizable in the first place. Thus, significant changes need to be made to the empirical practices in this field. To this end, we provided a cross-domain evaluation setup and tested several cyberbullying detection models, under a range of different representations to potentially overcome the limitations of the available data, and provide a fair, rigorous framework to facilitate direct model comparison for this task.

Additionally, we formed three hypotheses we would expect to find evidence for during these evaluations: \textbf{1}) the corpora are too small and heterogeneous to represent the strong variation in language-use for both bullying and neutral content across platforms accurately, \textbf{2}) the positive instances are biased, predominantly capturing toxicity, and no other dimensions of bullying, and finally \textbf{3}) crowdsourcing poses a resource to generate plausible cyberbullying events, and that can help expand the available data and improve the current models. 

We found evidence for all three hypotheses: previous cyberbullying models generalize poorly across domains, simple BoW baselines prove difficult to improve upon, there is considerable overlap between toxicity classification and cyberbullying detection, and crowdsourced data yields well-performing cyberbullying detection models. We believe that the results of Hypotheses \textbf{1}) and \textbf{2} in particular are principal hurdles that need to be tackled to advance this field of research. Furthermore, we showed that both leveraging training data from all openly available corpora, and shifting representations to include context meaningfully improves performance on the overall task. 
Therefore, we believe both should be considered as an evaluation point in future work. More so given that we show that these do not solve the existing limitations of the currently available corpora, and could therefore provide avenues for future research focusing on collecting (richer) data. Lastly, we show reproducibility of models that previously demonstrated state-of-the-art performance on this task to fail. We hope that the observations and contributions made in this paper can aid to improve rigor in future cyberbullying detection work.

\subsection*{Acknowledgements}

The work presented in this article was carried out in the framework of the AMiCA (IWT SBO-project 120007) project,
funded by the government agency for Innovation by Science and Technology (IWT). We would like to thank Prodromos Ninas, Kostas Stoitsas, and Alejandra Hern\'{a}ndez Rej\'{o}n for carrying out a range of trial experiments on this task, Bram Willemsen for helpful remarks, and in particular \'{A}kos K\'{a}d\'{a}r for the many discussions throughout all stages of the current work.

\bibliographystyle{acm}  
\bibliography{main.bib}

\end{document}